\documentclass[lettersize,journal,twoside]{IEEEtran}
\usepackage{amsmath,amsfonts}
\usepackage{algorithmic}
\usepackage{algorithm}
\floatname{algorithm}{\textcolor{black}{Algorithm}}
\usepackage{array}
\usepackage[caption=false,font=normalsize,labelfont=sf,textfont=sf]{subfig}
\usepackage{textcomp}
\usepackage{stfloats}
\usepackage{url}
\usepackage{verbatim}
\usepackage{graphicx}
\usepackage{cite}
\usepackage[most]{tcolorbox} 
\usepackage{enumitem}        
\usepackage{tabularx}  
\usepackage{amssymb}
\usepackage{booktabs}
\usepackage{siunitx}
\usepackage{multirow}
\usepackage{multirow}   
\usepackage{booktabs}   
\usepackage[table]{xcolor}
\usepackage{makecell}
\newcommand{\gcell}[1]{\cellcolor{black!8}#1}
\usepackage{pifont} 
\newcommand{\cmark}{\textcolor{green!60!black}{\normalsize\ding{51}}} 
\newcommand{\xmark}{\textcolor{red}{\normalsize\ding{55}}} 

\usepackage[hidelinks]{hyperref}
\usepackage{xcolor}
\usepackage{soul}
\definecolor{myblue}{RGB}{0,0,255}
\newcolumntype{C}[1]{>{\centering\arraybackslash}m{#1}}

\usepackage{etoolbox}
\AtBeginEnvironment{algorithm}{\color{black}}
\usepackage{orcidlink}
\AtBeginEnvironment{thebibliography}{%
  \let\origemph\emph
  \renewcommand{\emph}[1]{%
    \begingroup
      \def\etal{et~al.}%
      \edef\ARG{#1}%
      \ifx\ARG\etal
        \textup{#1}%
      \else
        \origemph{#1}%
      \fi
    \endgroup
  }%
}

\begin{document}
\bstctlcite{BSTcontrol} 

\title{VideoScout: Learning Agentic Active Exploration with Adaptive Reasoning Pacing for Long Video Understanding}

\author{Weixin Xu\,\orcidlink{0009-0003-4397-1472}, Zhenyu Yang\,\orcidlink{0009-0005-5298-0543}, Bing Wang\,\orcidlink{0009-0000-1018-6425}, Shengsheng Qian\,\orcidlink{0000-0001-9488-2208},~\IEEEmembership{Member,~IEEE}, Changsheng Xu\,\orcidlink{0000-0001-8343-9665},~\IEEEmembership{Fellow,~IEEE}
}

\markboth{PREPRINT, 2026}
{PREPRINT, 2026}


\maketitle
\begin{abstract}
Multimodal Large Language Models (MLLMs) have achieved remarkable progress on short video understanding yet remain limited on long videos due to the limited visual context window. Prevailing approaches rely on uniform frame sampling or recent coarse-to-fine agentic zooming, both of which struggle to localize sparse, decisive evidence in sufficiently long videos. We formulate long video understanding as a \textbf{Sequential Evidence Acquisition (SEA)} problem, in which an agent reads the video turn by turn along the temporal axis, deciding at each turn how fast to watch, what evidence to retain, when to revisit uncertain segments, and when to stop and answer. Inspired by this view, we propose \textbf{VideoScout}, a multi-turn reasoning agent that instantiates the SEA paradigm through adaptive reasoning pacing. Specifically, by dynamically controlling the viewing pace, VideoScout enables efficient traversal of long videos within a bounded visual context window, allowing the agent to access more video content while balancing content analysis depth with reading efficiency. To train VideoScout, we construct VideoScout-66K, a set of over 66K high-quality exploration turns from 10K answer-verified trajectories, and adopt a two-stage pipeline: cold-start supervised fine-tuning teaches the agent per-turn output format, while the Decoupled Clip and Dynamic sAmpling Policy Optimization (DAPO) algorithm performs trajectory-level reinforcement learning with a composite reward that jointly considers answer accuracy, output format compliance, and the temporal alignment between the agent's viewing progress and the teacher's answer timing measured by intersection-over-union (IoU). Extensive experiments on long video understanding and reasoning benchmarks demonstrate that our 7B model achieves strong performance compared with existing trained 7B agentic models.
\end{abstract}


\section{Introduction}
\IEEEPARstart{L}{ong} video understanding underpins a wide range of real-world applications, from video surveillance \textcolor{myblue}{\cite{chen2019distributed,tsakanikas2018video}} and instructional content analysis \textcolor{myblue}{\cite{miech2019howto100m,tang2019coin,zhong2023learning,shvetsova2024howtocaption}} to long-range film understanding\textcolor{myblue}{\cite{huang2020movienet,song2025moviechatplus,he2024ma}}. Recent Multimodal Large Language Models (MLLMs) have achieved strong performance on short video clips\textcolor{myblue}{\cite{Bai2025Qwen25VLTR,wang2025internvideo2,feng2026video,zhang2024long,li2025videochatrl}}. However, long videos introduce a fundamental bottleneck: a two-hour film at 24 fps contains more than 170,000 frames, and even when downsampled to 1 fps it still exceeds the visual context window of current 7B MLLMs by more than an order of magnitude\textcolor{myblue}{\cite{chen2024expanding,Bai2025Qwen25VLTR}}. As a result, the model can observe only a limited subset of frames, and the selected subset directly determines which evidence is available for answering the query. Long-video understanding therefore reduces to a question of how the model, under a bounded per-turn visual capacity, should acquire effective evidence from a video that vastly exceeds what it can see at once.

\begin{figure*}[!t]
\centering
\includegraphics[width=7.1in]{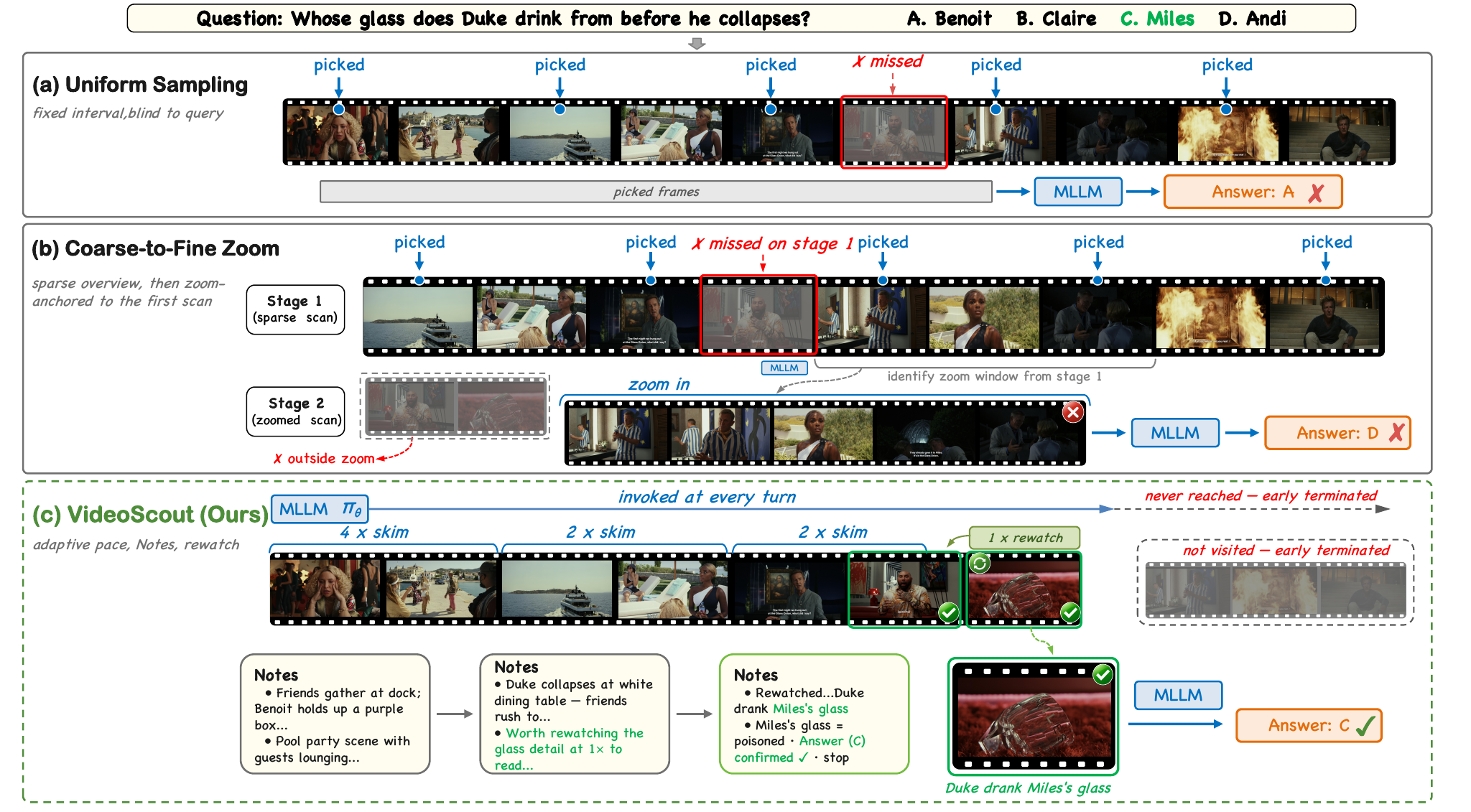}
\caption{VideoScout versus prior long video understanding paradigms on the same event-centric query. (a) Uniform sampling picks frames at fixed intervals and simply does not include the key frame. (b) Coarse-to-fine zoom anchors its zoom window on a sparse first-pass summary that has already missed the key frame, so the zoom window lands on the wrong interval and the evidence remains outside. (c) VideoScout consumes the video turn by turn, adapts its viewing pace over contiguous turn spans (bracketed $N{\times}$ labels), maintains evolving textual Notes, and rewatches an earlier turn from the current turn when new evidence calls for it, arriving at the correct answer.}
\label{fig_overview}
\end{figure*}

A first line of work addresses this problem with \textit{predetermined observation strategies}. Uniform frame sampling\textcolor{myblue}{\cite{zhang2024llava,li2024llava,lin2024video,li2025videochat,ataallah2024minigpt4,Bai2025Qwen25VLTR,wang2024uni,wu2023stmixer,fei2024enhancing}} and query-conditioned keyframe selection\textcolor{myblue}{\cite{tang2025adaptive,liang2024keyvideollm,yu2025frame,wang2025videotree,yu2023self,wei2026cfvbench,liu2026moment,hu2025cos}} commit to a frame set before reasoning begins, so once decisive evidence is excluded, no subsequent reasoning can recover it, as illustrated in Fig.~\textcolor{myblue}{\ref{fig_overview}}(a). Coarse-to-fine agentic methods such as LongVT\textcolor{myblue}{\cite{yang2026longvt}} and VideoTemp-o3\textcolor{myblue}{\cite{liu2026videotemp}} introduce multiple turns and iteratively zoom into informative temporal windows identified from an initial low-frame-rate overview\textcolor{myblue}{\cite{ma2025drvideo,yang2025vca}}, but the zoom selection remains anchored to a one-shot global scan that, on sufficiently long videos, is itself too sparse to indicate where the decisive evidence lies, as illustrated in Fig.~\textcolor{myblue}{\ref{fig_overview}}(b). In both cases, the model never actively seeks out the sparse evidence scattered across the video; it can only reason over whatever an up-front observation plan happens to surface. This motivates our \textbf{Core Challenge 1}: \textit{how can we build an agentic active exploration strategy for long videos, where decisive evidence is sparse and difficult to gather and reason over?}

\IEEEpubidadjcol

Yet active exploration alone is not enough. Long videos contain a substantial amount of redundant or irrelevant content, and reasoning indiscriminately over every chunk is neither feasible nor desirable, since many segments simply do not warrant careful reasoning. Existing efforts attempt to address this redundancy by compressing long videos into memory representations\textcolor{myblue}{\cite{he2024ma,song2024moviechat,shu2024videoxl}} or by pre-selecting query-relevant frames\textcolor{myblue}{\cite{tang2025adaptive,wang2025videotree}}, but in both cases the allocation of visual capacity is decided once and frozen, with no way to dynamically refine it as understanding of which parts are truly relevant to the query develops over time. This motivates our \textbf{Core Challenge 2}: \textit{under a bounded visual capacity, how can such an agent achieve efficient observation and reasoning by concentrating its effort on segments worth examining?}

To address both challenges, we propose \textbf{VideoScout}, a trained multi-turn video agent that interacts with the video turn by turn rather than committing to a one-shot observation plan. For \textbf{Challenge 1}, VideoScout introduces cross-turn textual notes coupled with a \texttt{rewatch} action to realize agentic active exploration: at each turn the agent observes only a short segment and reasons over its accumulated notes, then may either continue forward or revisit a previously fast-forwarded chunk for closer inspection; the notes carry query-relevant evidence across turns, so sparse decisive evidence is gathered iteratively and every new observation is conditioned on what the agent has already understood. For \textbf{Challenge 2}, VideoScout equips the agent with an adaptive \texttt{speed} action that realizes efficient observation and reasoning under bounded visual capacity: the agent fast-forwards through uninformative passages at $2\times$ or $4\times$ and slows down to $1\times$ on segments worth examining, concentrating its limited capacity exactly where it matters; a final \texttt{answer} action commits the rollout once enough evidence has been gathered, avoiding indiscriminate reasoning over every chunk. Fig.~\textcolor{myblue}{\ref{fig_overview}}(c) illustrates VideoScout on the same event-centric query.

We train VideoScout on top of Qwen2.5-VL-7B\textcolor{myblue}{\cite{Bai2025Qwen25VLTR}}  with a two-stage pipeline tailored to multi-turn agentic exploration. We first construct \textbf{VideoScout-66K}, a dataset of over $66$K verified exploration turns from $10$K answer-verified trajectories, and perform turn-level cold-start supervised fine-tuning (SFT) to teach the agent the output format and per-turn action selection behavior. We then apply the Decoupled Clip and Dynamic sAmpling Policy Optimization (DAPO)\textcolor{myblue}{\cite{yu2026dapo}} algorithm at the trajectory level with a composite reward that jointly considers answer accuracy, output format compliance, and temporal alignment with teacher viewing behavior measured by intersection-over-union, thereby aligning the optimization objective with the complete multi-turn reasoning process rather than isolated individual turns.

At a higher level, the design of VideoScout reflects a more fundamental shift in how long-video understanding should be approached: from passively encoding a fixed set of frames to actively \textit{reading} a long video as a human would. A human viewer does not commit to a fixed observation strategy in advance; instead, she skims irrelevant passages, slows down at important moments, takes notes of salient cues, and occasionally revisits segments that were passed too quickly. Each action depends on what has already been observed, allowing the observation strategy to continuously evolve together with the reasoning process. We formalize this human-like reading as a paradigm: long-video question answering under bounded visual context is a \textbf{Sequential Evidence Acquisition (SEA)} problem, in which the agent reads the video turn by turn along the temporal axis, deciding at each turn how fast to watch, what evidence to retain, when to revisit uncertain segments, and when to stop and answer, turning the question of \textit{how to encode a fixed video} into \textit{how to allocate limited observations over time}.

In summary, our contributions are as follows:
\begin{itemize}
    \item We formulate long-video question answering under bounded visual context as a turn-by-turn temporal reading process, in which an agent advances through the video sequentially in time and at each turn decides how fast to watch, what evidence to retain, when to revisit uncertain segments, and when to stop and answer, analogous to how a human reads a long passage. We refer to this formulation as the \textbf{Sequential Evidence Acquisition (SEA)} problem.

    \item We introduce \textbf{VideoScout}, a trained multi-turn video agent that implements SEA through adaptive-speed observation, compact textual notes, and local rewatch actions. By coupling observation and reasoning in a turn-by-turn process, VideoScout can trade off temporal coverage and visual detail, enabling efficient traversal of long videos while preserving the ability to inspect sparse decisive evidence.

    \item We construct \textbf{VideoScout-66K} comprising over 66K verified exploration turns from 10K answer-verified trajectories, and develop a two-stage training pipeline in which turn-level cold-start SFT learns the output format and per-turn action behavior, while trajectory-level DAPO with a composite reward over accuracy, format, and timing IoU aligns the optimization objective with the complete multi-turn reasoning process.

    \item Extensive experiments on long video understanding and reasoning benchmarks demonstrate that our 7B agent achieves strong performance compared with existing trained 7B agentic models.
\end{itemize}

\section{Related Work}
\subsection{Long Video Understanding}
Early approaches to long video understanding rely on uniform frame sampling\textcolor{myblue}{\cite{zhang2024llava,li2024llava,lin2024video,shi2025mavors,zhao2025efficient,azad2025hierarq,wang2025test}} or query-conditioned keyframe selection\textcolor{myblue}{\cite{xu2026long,tan2026msjoe,tang2026tspo,yu2023self,zhang2026one,hu2025unified,chang2026ca2st,zhao2021reconstructive}}, which commit to a fixed set of frames before reasoning and therefore often miss sparse yet decisive evidence. For instance, query-conditioned keyframe selection methods such as Adaptive Keyframe Sampling (AKS)\textcolor{myblue}{\cite{tang2025adaptive}} typically select keyframes based on semantic relevance to the query through image-text matching scores, which may still overlook temporally dispersed evidence that requires multi-step reasoning to connect. Another line of work augments MLLMs with running memory. Methods such as MA-LMM\textcolor{myblue}{\cite{he2024ma}}, MovieChat\textcolor{myblue}{\cite{song2024moviechat}}, Flash-VStream\textcolor{myblue}{\cite{zhang2024flashvstream}}, LongVU\textcolor{myblue}{\cite{shen2024longvu}}, and Video-XL\textcolor{myblue}{\cite{shu2024videoxl}} compress videos into continuously updated memory representations, but fine-grained visual cues may be lost once compressed. More recently, training-free agentic pipelines\textcolor{myblue}{\cite{fan2024videoagent,shang2024traveler,zhang2026deep,zhang2024omagent,llovi,morevqa}} employ powerful proprietary large language models to iteratively query clip-level visual tools for reasoning. For example, Deep Video Discovery\textcolor{myblue}{\cite{zhang2026deep}} treats long video understanding as agentic search with tool use for temporal retrieval and object tracking. Although effective, their performance is tightly coupled to closed-source backbones, making them difficult to reproduce, deploy, or directly compare with trained open-source models. In contrast, our approach trains a self-contained 7B agent whose observation policy is jointly optimized together with the reasoning process in an end-to-end manner.
\subsection{Tool-Augmented Video Agents with Reinforcement Learning}
Reinforcement learning, originally adopted to align language models with human preferences\textcolor{myblue}{\cite{ouyang2022instructgpt,rafailov2023direct,ethayarajh2024kto,meng2024simpo,yuan2024self}}, has become a key paradigm for enhancing the reasoning ability of multimodal large language models\textcolor{myblue}{\cite{huang2025vision,liu2025visual,yu2024rlhf}}, with Group Relative Policy Optimization (GRPO)\textcolor{myblue}{\cite{shao2024deepseekmath}} emerging as a particularly effective optimization algorithm. The success of this paradigm is further demonstrated by DeepSeek-R1\textcolor{myblue}{\cite{guo2025deepseekr1}}, which scales reasoning RL to elicit chain-of-thought capabilities at scale. Building upon this paradigm, recent works\textcolor{myblue}{\cite{li2025videochatrl,wang2026videorft,park2026deepvideo,chen2026vragent,meng2025open}} extend GRPO-style post-training from text to video, achieving strong improvements on video question answering and temporal grounding tasks. For example, Video-R1\textcolor{myblue}{\cite{feng2026video}} proposes T-GRPO to explicitly encourage temporal reasoning during RL post-training. However, these methods largely remain single-turn: the model generates both reasoning traces and final answers in a single pass over passively provided visual inputs, without actively interacting with the video during inference. A parallel line of research introduces agentic pipelines through coarse-to-fine temporal zoom strategies\textcolor{myblue}{\cite{rasheed2025video,xie2025video,liu2025videomind,liu2026videotemp}}. For example, LongVT\textcolor{myblue}{\cite{yang2026longvt}} first inspects a sparse low-frame-rate overview and then iteratively zooms into informative temporal windows identified from the overview. Although substantially more adaptive than single-turn baselines, its exploration strategy remains anchored to an initial global scan, whose reliability degrades as video duration increases. Our work advances this direction by formulating long video understanding as a genuine multi-turn reasoning problem under bounded visual context, and by optimizing the agent's observation policy using the DAPO algorithm\textcolor{myblue}{\cite{yu2026dapo}} with a composite reward over answer accuracy, output format compliance, and temporal alignment via IoU-based timing rewards.

\begin{figure*}[!t]
\centering
\includegraphics[width=7.1in]{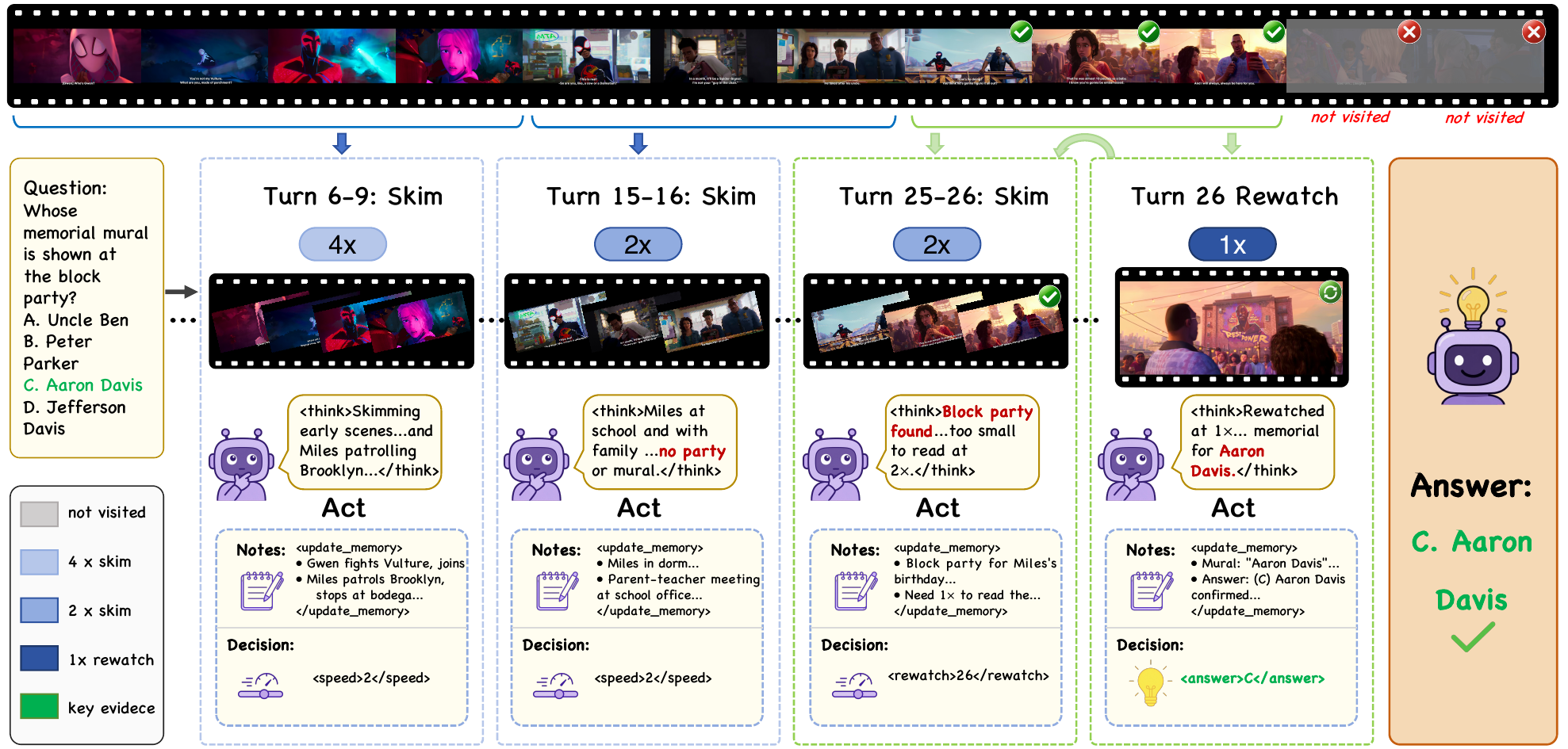}
\caption{Overview of the VideoScout framework for long-video reasoning. Given a long video and a query, the video is first split into multiple fixed-duration chunks. The agent then enters an iterative multi-turn reasoning process, where at each turn it receives the current visual clip and the textual memory from the previous turn, generates a reasoning trace, and selects one action from three options: \texttt{speed}$(s)$ adjusts the playback pace to advance quickly or slow down, \texttt{rewatch}$(k)$ requests a closer re-examination of a specific chunk for more detail, and \texttt{answer}$(\hat{y})$ commits to the final answer, terminating the rollout.}
\label{fig_pipeline}
\end{figure*}

\section{Methodology}
This section details how VideoScout instantiates the Sequential Evidence Acquisition (SEA) paradigm as a trained multi-turn agent. We first present the task formulation and agent behavior, then describe the data generation pipeline that yields VideoScout-66K, followed by the cold-start supervised fine-tuning and trajectory-level reinforcement learning stages.

\subsection{Overview}
\textit{1) Task formulation.}
Given a long video $V$ and a query $q$, we split $V$ into $N$ contiguous chunks $\{c_1, c_2, \ldots, c_N\}$ of fixed duration, and let an agent policy $\pi_\theta$ interact with the video turn by turn until it emits an $\texttt{answer}$ action. At each turn $t$, the agent observes only the current visual clip $v_t$ and the textual memory $m_{t-1}$ carried from the previous turn, and samples a structured output
\begin{equation}
  \bigl(r_t,\, m_t,\, a_t\bigr) \;\sim\; \pi_\theta\!\left(\cdot \mid v_t,\, m_{t-1},\, q\right),
  \label{eq:agent_step}
\end{equation}
consisting of a reasoning trace $r_t$, an optionally updated textual memory $m_t$, and exactly one action $a_t$ drawn from
\begin{equation}
  a_t \;\in\; \bigl\{\texttt{speed}(s),\; \texttt{rewatch}(k),\; \texttt{answer}(\hat{y})\bigr\},
  \label{eq:action_space}
\end{equation}
where $s \in \{1{\times}, 2{\times}, 4{\times}\}$ controls the pace of the next observation, $\texttt{rewatch}(k)$ requests a closer re-examination of a chunk $c_k$ drawn from the current observation window, and $\texttt{answer}(\hat{y})$ terminates the rollout.

\textit{2) Agent behavior.}
The clip $v_t$ and its pace $s$ are coupled through a fixed per-turn visual token budget: at $\texttt{speed}(1{\times})$, $v_t$ covers a single chunk densely sampled at the highest per-frame resolution the budget allows, while higher speeds cover proportionally more chunks under the same budget, trading frame density and per-frame resolution for temporal coverage. The textual memory $m_t$ has a fixed word budget $L_m$; at each turn the agent may update the memory by writing a new complete version of $m_t$, or leave $m_t = m_{t-1}$ when nothing new is relevant to $q$. Because an update carries the entire memory forward rather than a diff, the agent must actively decide at each update which prior observations to keep, rephrase, or drop. The $\texttt{rewatch}(k)$ action is available only when $v_t$ spans multiple chunks, i.e., when the current pace is $2{\times}$ or $4{\times}$; the chosen index $k$ must refer to a chunk inside the current window, and the next turn is dedicated to re-examining $c_k$ alone at $1{\times}$. Forward progression then automatically resumes from the chunk immediately after the original fast-forward window, so rewatching recovers detail missed during fast-forwarding without rewinding the agent's overall progress through $V$. The $\texttt{answer}(\hat{y})$ action is emitted when the accumulated memory and observations are deemed sufficient to respond to $q$. Beyond the notation defined above, we denote the current observation window by $W$ and use a forward pointer $p$ to mark the next unseen chunk, which advances only on forward observations so that the agent moves through the video monotonically apart from on-demand rewatching. Fig.~\textcolor{myblue}{\ref{fig_pipeline}} illustrates one such rollout, and we summarize this active exploration process in Algorithm~\textcolor{myblue}{\ref{alg:active_exploration}}.

\begin{algorithm}[t]
\caption{Active Exploration Reasoning of VideoScout}
\label{alg:active_exploration}
\begin{algorithmic}[1]
\REQUIRE video $V$, query $q$, policy $\pi_\theta$
\ENSURE predicted answer $\hat{y}$
\STATE split $V$ into contiguous chunks $\{c_1, c_2, \ldots, c_N\}$
\STATE \textit{// Initialize the exploration state}
\STATE turn $t \leftarrow 1$, forward pointer $p \leftarrow 1$, speed $s \leftarrow 1{\times}$
\STATE memory $m_0 \leftarrow \varnothing$, no pending rewatch
\WHILE{the agent has not committed to an answer}
  \STATE \textit{// Observation stage: form the clip for this turn}
  \IF{a rewatch of $c_k$ is pending}
    \STATE $v_t \leftarrow$ re-sample $c_k$ at $1{\times}$ for a closer look
    \STATE clear the pending rewatch
  \ELSE
    \STATE $W \leftarrow$ chunks from $c_p$ covered under speed $s$
    \STATE $v_t \leftarrow$ sample $W$ under speed $s$
    \STATE $p \leftarrow p + |W|$ \quad\textit{// advance the forward pointer}
  \ENDIF
  \STATE \textit{// Reasoning and action stage}
  \STATE $(r_t,\, m_t,\, a_t) \sim \pi_\theta(\cdot \mid v_t,\, m_{t-1},\, q)$
  \IF{$a_t = \texttt{answer}(\hat{y})$}
    \STATE \textit{// Commit: evidence is sufficient}
    \RETURN $\hat{y}$
  \ELSIF{$a_t = \texttt{rewatch}(k)$ and $s \neq 1{\times}$ and $c_k \in W$}
    \STATE \textit{// Revisit: re-examine a skimmed chunk next turn}
    \STATE mark $c_k$ as pending rewatch
  \ELSIF{$a_t = \texttt{speed}(s')$}
    \STATE \textit{// Re-pace: skim faster or slow down}
    \STATE $s \leftarrow s'$
  \ENDIF
  \STATE $t \leftarrow t + 1$
\ENDWHILE
\STATE \textbf{Note:} the memory $m_t$ is the only state carried across turns; the forward pointer $p$ advances solely on forward observations, so the agent traverses $V$ monotonically except for the on-demand rewatch.
\end{algorithmic}
\end{algorithm}

\textit{3) Scalability.}
Both $v_t$ and $m_{t-1}$ are of bounded size by construction, so the per-turn inference cost of $\pi_\theta$ is constant in the video length $|V|$. The total cost of a rollout therefore scales only with the number of turns taken, which in turn depends on how aggressively the agent fast-forwards rather than on the underlying video length.

\begin{figure}[!t]
\centering
\includegraphics[width=3.4in]{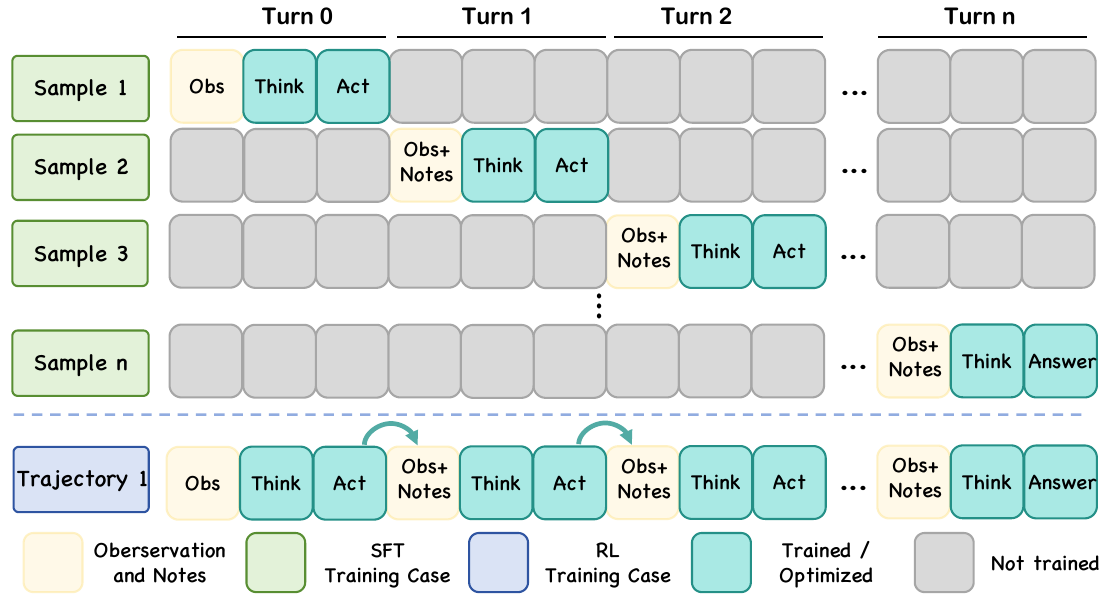}
\caption{Comparison between the SFT and RL training schemes. SFT optimizes individual turns independently at the turn level, where each training case corresponds to a single turn of a trajectory, whereas RL optimizes the entire multi-turn trajectory as a whole at the trajectory level.}
\label{fig_train_scheme}
\end{figure}

\textit{4) Training overview.}
We train VideoScout in two stages that operate on the same trajectories at different granularities, as illustrated in Fig.~{\textcolor{myblue}{\ref{fig_train_scheme}}}. The first stage is a cold-start supervised fine-tuning that treats each turn of a trajectory as an independent training example and optimizes it at the turn level, teaching the agent the output format and per-turn action behavior. The second stage is reinforcement learning that optimizes the entire multi-turn trajectory as a whole at the trajectory level, aligning the objective with the complete exploration process rather than isolated turns. This two-granularity view motivates our data construction, which produces both turn-level supervision and complete trajectories, as described next.

\subsection{Data Generation Pipeline}

\begin{figure*}[!t]
\centering
\includegraphics[width=7.1in]{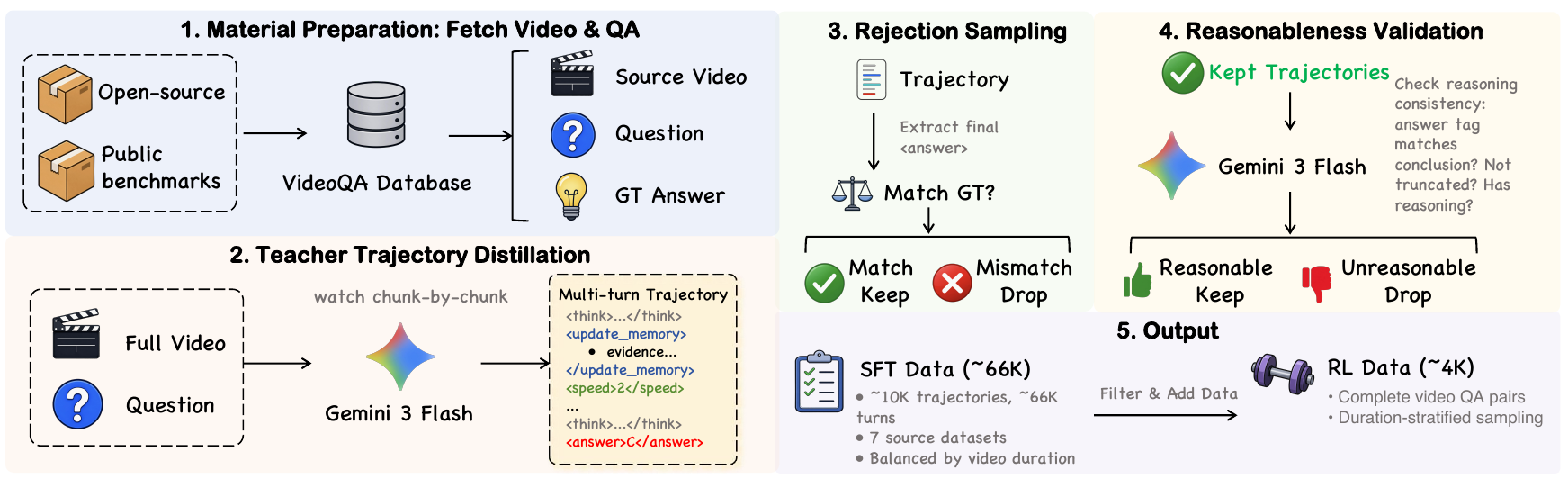}
\caption{The five-stage data generation pipeline of VideoScout-66K: Material Preparation, Teacher Trajectory Distillation, Rejection Sampling, Reasonableness Validation, and Output.}
\label{fig_data_pipeline}
\end{figure*}

To cold-start the multi-turn reasoning agent, we construct the VideoScout-66K dataset, a collection of over 66K verified exploration turns from 10K answer-verified trajectories for long-video multiple-choice question answering. We build this dataset through a five-stage data generation pipeline, as shown in Fig.~{\textcolor{myblue}{\ref{fig_data_pipeline}}}.

\textit{1) Material Preparation.} We curate training queries from publicly available long-video MCQ benchmarks, including VideoMarathon\textcolor{myblue}{\cite{lin2026unleashing}}, ScaleLong\textcolor{myblue}{\cite{ma2025scalelong}}, LongVideo-Reason\textcolor{myblue}{\cite{chen2026scaling}}, and TVQA-Long\textcolor{myblue}{\cite{ataallah2024goldfish}}. Each query is accompanied by its source video and ground-truth answer.

As shown in Tab.~{\ref{tab:video_sampling}}, the visual sampling configuration varies across different viewing modes. Tab.~{\ref{tab:chunk_strategy}} shows the chunk duration, which is automatically determined based on the total video length. Both training and inference use the same chunking and sampling logic.

\textit{2) Teacher Trajectory Distillation.} We employ Gemini 3 Flash\textcolor{myblue}{\cite{gemini3flash}} as the teacher model. The teacher receives the complete VideoScout system prompt, including the action space, the Think-Act format, and the memory mechanism, but without the ground-truth answer. It produces a multi-turn exploration trajectory, deciding at each step how to watch and when to answer while maintaining a textual memory.

\textit{3) Rejection Sampling.} Some generated trajectories may produce incorrect final answers. We resample a trajectory up to eight times until the teacher's final answer matches the ground truth, and discard the question if no correct trajectory is obtained within this budget. Only trajectories with correct answers are retained.

\textit{4) Reasonableness Validation.} Beyond answer correctness, we perform additional filtering to ensure the trajectory is reasonable: reasoning consistency is checked, the answer tag matches the trajectory conclusion, no premature truncation is detected, and the trajectory contains meaningful reasoning traces.

\begin{figure}[!t]
\centering
\includegraphics[width=3.4in]{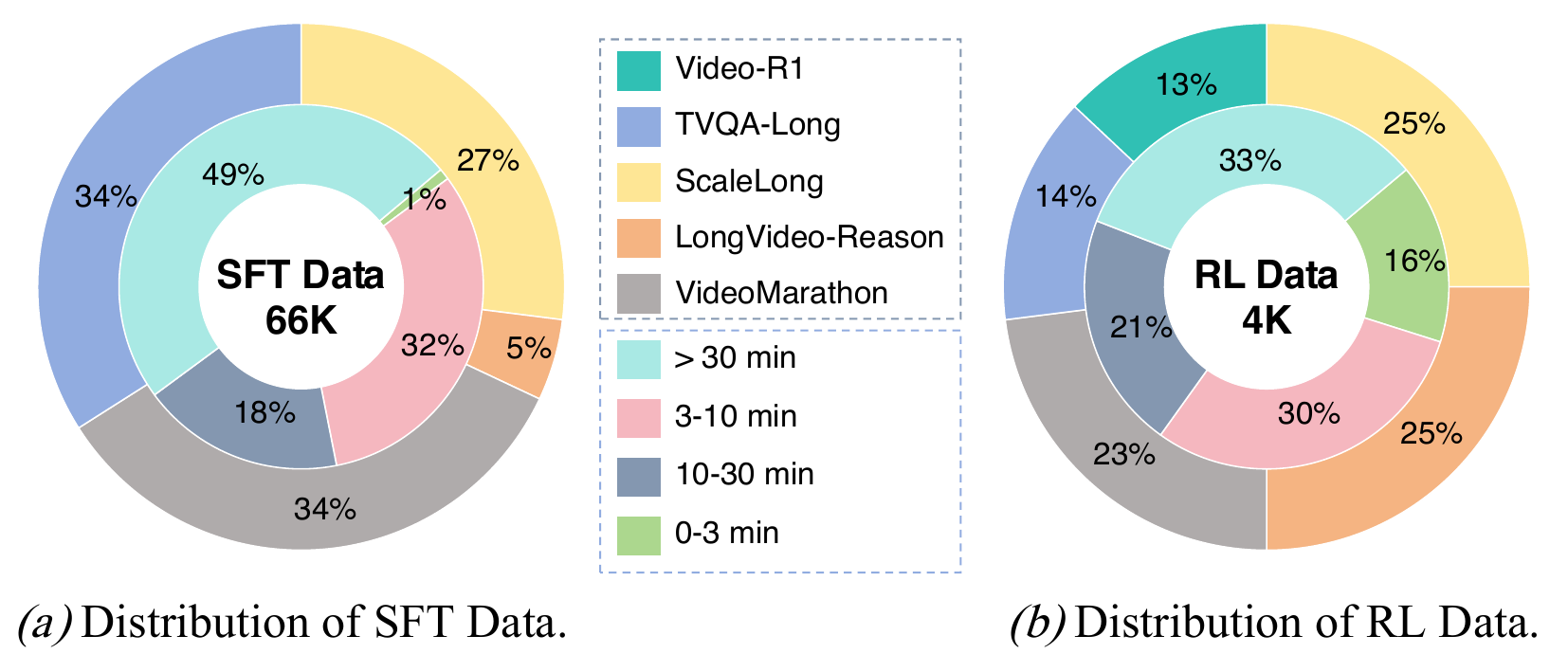}
\caption{Distribution of the training data. (a) The SFT data (66K turns) and (b) the RL data (4K prompts). The outer ring shows the composition of data sources, and the inner ring shows the video duration distribution.}
\label{fig_data_distribution}
\end{figure}

\textit{5) Output.} After filtering, the pipeline yields approximately $10$K verified trajectories comprising over $66$K individual turns, with a duration distribution skewed toward long videos as shown in Fig.~{\textcolor{myblue}{\ref{fig_data_distribution}}(a)}: approximately $49\%$ of samples exceed $30$ minutes, $27\%$ fall within $3$-$10$ minutes, $23\%$ are under $3$ minutes, and $1\%$ span $10$-$30$ minutes. These trajectories are used for supervised fine-tuning. For reinforcement learning, we first filter high-quality queries from validated trajectories, then supplement with additional Video-R1\textcolor{myblue}{\cite{feng2026video}} samples and re-balance the duration distribution to form a diverse pool of roughly $4$K prompts as shown in Fig.~{\textcolor{myblue}{\ref{fig_data_distribution}}}(b). Compared to the SFT data, the RL pool exhibits a more uniform duration distribution with $33\%$ exceeding $30$ min, $30\%$ within $3$-$10$ min, $25\%$ spanning $10$-$30$ min, and $12\%$ under $3$ min, which ensures that the policy receives balanced exploration signals across short, medium, and long video scenarios.

\subsection{Cold-Start Supervised Fine-Tuning}
\textit{Single-turn supervised fine-tuning.}
For supervised fine-tuning, we treat every turn of every trajectory as an independent training example: the input consists of the query $q$, the current visual clip $v_t$, and the inherited memory $m_{t-1}$, and the target is the structured output $(r_t, m_t, a_t)$ defined by Eq.~{\eqref{eq:agent_step}}. We optimize the standard token-level cross-entropy loss on the agent's output only, masking out the visual clip and the user prompt. This single-turn imitation teaches the model the Think-Act format, the legal action space of Eq.~{\eqref{eq:action_space}}, and a first pass over how to trade off pace, memory updates, and rewatching based on the currently observed content. The agent is initialized from Qwen2.5-VL-7B-Instruct\textcolor{myblue}{\cite{Bai2025Qwen25VLTR}} with all parameters fine-tuned.

\subsection{Trajectory-Level Reinforcement Learning}
\textit{1) Multi-turn rollout.}
The SFT-initialized agent produces well-formed trajectories but receives no direct signal that rewards accurate and temporally efficient exploration. We therefore optimize the full multi-turn rollout with Decoupled Clip and Dynamic sAmpling Policy Optimization (DAPO)\textcolor{myblue}{\cite{yu2026dapo}} on roughly $4$K prompts that are collected through filtering validated trajectories and supplementing with additional data sources, sampled in balance across short, medium, and long video durations as well as diverse data sources. For each $(V, q)$, we sample a group of $G$ independent trajectories $\{\tau_i\}_{i=1}^G$ from $\pi_\theta$, each unrolled until an $\texttt{answer}$ is emitted. For trajectory $\tau_i$ with terminal reward $R(\tau_i)$, the group-relative advantage is
\begin{equation}
  A_i \;=\; \frac{R(\tau_i) - \tfrac{1}{G}\sum_j R(\tau_j)}{\mathrm{std}_j R(\tau_j)},
  \label{eq:advantage}
\end{equation}
and we optimize the decoupled clipped surrogate objective over the agent's output tokens along every trajectory, with separate upper and lower clipping bounds to encourage exploration:
\begin{equation}
\begin{aligned}
\mathcal{L}_{\text{DAPO}}(\theta)
=&
\mathbb{E}_{(V,q)\sim\mathcal{D},\;
\{\tau_i\}_{i=1}^{G}\sim\pi_{\theta_{\text{old}}}}\\
&\biggl[
\frac{1}{
\sum_{i=1}^{G}
\sum_{j=1}^{n_i}
|o_{i,j}|
}
\sum_{i=1}^{G}
\sum_{j=1}^{n_i}
\sum_{t=1}^{|o_{i,j}|}\\
&\left(
C_{i,j,t}
-
\beta D_{\mathrm{KL}}
\left(
\pi_\theta \,\|\, \pi_{\mathrm{ref}}
\right)
\right)
\biggr],
\end{aligned}
\label{eq:dapo_loss}
\end{equation}
where $n_i$ is the number of output turns in trajectory $\tau_i$, $o_{i,j}$ denotes the output sequence of turn $j$, $o_{i,j,t}$ its $t$-th token, and the per-token clipped surrogate is
\begin{equation}
\begin{aligned}
C_{i,j,t} \;=\;& \min\!\Bigl( r_{i,j,t}(\theta) \, A_i,\; \\ &\operatorname{clip}\!\bigl( r_{i,j,t}(\theta),\; 1 - \epsilon_{\text{low}},\; 1 + \epsilon_{\text{high}} \bigr) \, A_i \Bigr),
\end{aligned}
\label{eq:dapo_clip}
\end{equation}
where the importance sampling weight is
\begin{equation}
\begin{aligned}
r_{i,j,t}(\theta) = \frac{\pi_\theta(o_{i,j,t} \mid q, o_{i,<t})}{\pi_{\theta_{\text{old}}}(o_{i,j,t} \mid q, o_{i,<t})}.
\end{aligned}
\label{eq:dapo_r}
\end{equation}
Here $A_i$ is the group-relative advantage from Eq.~{\eqref{eq:advantage}}, shared by all output tokens of trajectory $\tau_i$, and $\epsilon_{\text{low}}$ and $\epsilon_{\text{high}}$ are independent lower and upper clipping thresholds that decouple the clipping bounds to allow more aggressive exploration. Following DAPO, we additionally adopt dynamic sampling, which discards prompts whose sampled group of trajectories is entirely correct or entirely incorrect, as such groups yield zero advantage and no gradient, and continues sampling until the batch is filled with informative prompts.

\begingroup
\setlength{\tabcolsep}{3.5pt}
\begin{table}[t]
\centering
\caption{VideoScout video sampling configuration. Frame count, resolution, and coverage for different viewing modes.}
\label{tab:video_sampling}
\begin{tabular}{c|c|c|c}
\toprule
\textbf{Speed} & \textbf{Frames} & \textbf{Resolution} & \textbf{Covered Chunks} \\
\midrule
Normal ($1\times$) & 20 & $448\times448$ & 1 \\
$2\times$ Fast-forward & 16 & $448\times448$ & 2 \\
$4\times$ Fast-forward & 12 & $316\times316$ & 4 \\
Rewatch & 20 & $448\times448$ & 1 \\
\bottomrule
\end{tabular}
\end{table}
\endgroup

\begingroup
\setlength{\tabcolsep}{5pt}
\begin{table}[t]
\centering
\caption{VideoScout chunk segmentation strategy. Chunk duration based on video length.}
\label{tab:chunk_strategy}
\begin{tabular}{c|c}
\toprule
\textbf{Video Duration} & \textbf{Chunk Duration} \\
\midrule
$<$ 30s & 10s \\
30s -- 2min & 15s \\
2min -- 5min & 20s \\
5min -- 15min & 30s \\
$\ge$ 15min & 45s \\
\bottomrule
\end{tabular}
\end{table}
\endgroup

\textit{2) Composite reward.}
The terminal reward $R(\tau)$ jointly captures answer quality, output validity, and temporal alignment between the agent's viewing progress and the teacher's answer timing:
\begin{equation}
  R(\tau) \;=\; r_{\text{acc}}(\tau) \;+\; r_{\text{fmt}}(\tau) \;+\; r_{\text{timing}}(\tau),
  \label{eq:reward}
\end{equation}
where $r_{\text{acc}}(\tau)\in\{0,1\}$ is an accuracy reward obtained by matching the predicted answer $\hat{y}$ against the ground truth, and $r_{\text{fmt}}(\tau)\in\{0,0.3\}$ rewards well-formed outputs with valid $\texttt{answer}$ tags and no malformed turns. The timing reward incentivizes the agent to answer at a viewing progress similar to the teacher trajectory, which is critical for ensuring that the agent makes its answer at an appropriate moment during video exploration. Letting $M$ denote the agent's forward progress when it commits to an answer, measured as the number of chunks it has advanced through, and $T$ the chunk index at which the teacher answered, we compute the Intersection over Union (IoU) between the agent's viewing interval $A=[0, M]$ and the teacher's viewing interval $B=[0, T]$:
\begin{equation}
  \text{IoU} \;=\; \frac{|A \cap B|}{|A \cup B|} \;=\; \frac{\min(M, T)}{\max(M, T)},
  \label{eq:iou}
\end{equation}
and then scale it to obtain the timing reward:
\begin{equation}
  r_{\text{timing}}(\tau) \;=\; \text{IoU} \cdot 0.5 \cdot \mathbb{I}\!\left[r_{\text{acc}}(\tau) = 1\right],
  \label{eq:timing}
\end{equation}
The IoU measures the overlap between the agent's and teacher's viewing intervals, both starting from the first chunk. Crucially, $r_{\text{timing}}$ is awarded only when the answer is correct, preventing the policy from trading accuracy for timing alignment.

\section{Experiments}

\subsection{Experimental Setup}
\textit{1) Implementation Details.}
We initialize our model from Qwen2.5-VL-7B-Instruct\textcolor{myblue}{\cite{Bai2025Qwen25VLTR}} for its strong foundational capabilities and multimodal comprehension. Our training consists of two stages. In the cold-start supervised fine-tuning stage, we train on the VideoScout-66K dataset comprising over $66$K exploration turns from $10$K answer-verified trajectories. We set the learning rate to $1.5\times10^{-5}$ with a cosine scheduler and $5\%$ warmup, a global batch size of $16$, and train for $3$ epochs on $8$ NVIDIA H800 GPUs. We adopt packing\textcolor{myblue}{\cite{krell2021efficient}} with a maximum sequence length of $16{,}384$ tokens. For video input, sampling is performed as shown in Tab.~{\ref{tab:video_sampling}}.

In the reinforcement learning stage, we apply DAPO\textcolor{myblue}{\cite{yu2026dapo}}  optimization over the trajectory-level rollout with a group size of $8$. We use a learning rate of $5\times10^{-6}$, a KL regularization coefficient of $0.01$ with low-variance KL estimation\textcolor{myblue}{\cite{shao2024deepseekmath}}, and decoupled clipping bounds of $\epsilon_{\text{low}}=0.20$ and $\epsilon_{\text{high}}=0.28$. The RL training also runs for $1$ epoch on $8$ NVIDIA H800 GPUs, with a maximum prompt length of $36{,}000$ tokens and a maximum response length of $4{,}096$ tokens.

\textit{2) Benchmarks.}
We evaluate VideoScout on nine benchmarks organized into two categories: long video understanding and reasoning benchmarks that assess holistic understanding and reasoning over long videos, including perception, temporal relation modeling, causal reasoning, and multi-step reasoning, and cross-domain video understanding benchmarks that evaluate generalization across diverse knowledge domains.

\textbf{Long video understanding and reasoning benchmarks.} This category comprises four benchmarks that primarily assess the model's holistic understanding and reasoning over long videos, covering perception, memory, temporal relation modeling, causal reasoning, and multi-step reasoning. MLVU\textcolor{myblue}{\cite{zhou2025mlvu}} contains videos ranging from 10 minutes to 2 hours with questions that require multi-dimensional reasoning over spatial, temporal, and causal cues; we report results on its held-out test set (MLVU-test). LVBench\textcolor{myblue}{\cite{wang2025lvbench}} features videos up to 2 hours with diverse question types including counting, spatial relation, temporal ordering, action recognition, and key information retrieval. Video-MME\textcolor{myblue}{\cite{fu2025video}} features videos from 6 to 60 minutes and covers a broad spectrum of tasks ranging from perception and recognition to temporal reasoning and information synthesis; we adopt the subtitle-free setting in our evaluation. MINERVA\textcolor{myblue}{\cite{nagrani2025minerva}} focuses on complex multi-step reasoning over long videos, requiring the model to integrate evidence across multiple temporal windows.

\textbf{Cross-domain video understanding benchmarks.} This category comprises five benchmarks that evaluate generalization across diverse knowledge domains, including multi-disciplinary knowledge, scientific videos, world commonsense, and visual perception. VideoMMMU\textcolor{myblue}{\cite{hu2025video}} requires multi-step reasoning over multi-disciplinary video content, with sub-tasks including comprehension, adaptation, and perception. SciVideoBench\textcolor{myblue}{\cite{deng2025scivideobench}} evaluates scientific video understanding that requires domain-specific knowledge. WorldSense\textcolor{myblue}{\cite{hong2025worldsense}} tests everyday video comprehension grounded in world commonsense; since our model takes only visual input, we evaluate on its vision-only setting without audio. VSI-Bench\textcolor{myblue}{\cite{yang2025thinking}} focuses on visual-spatial perception and integration. MMVU\textcolor{myblue}{\cite{zhao2025mmvu}} assesses integrated perception-and-reasoning capabilities across diverse video types; we evaluate on its multiple-choice subset (MMVU-mc).

\textit{3) Baseline Methods.}
We compare VideoScout against two groups of baselines: proprietary MLLMs and open-source MLLMs. Within each group, we further distinguish between standard single-pass models and agentic frameworks that perform iterative, tool-augmented exploration over the video.

\textbf{Proprietary MLLMs.} We include three representative closed-source models: GPT-4o\textcolor{myblue}{\cite{hurst2024gpt}}, Gemini-1.5-Pro\textcolor{myblue}{\cite{team2024gemini}}, and Gemini-2.5-Pro\textcolor{myblue}{\cite{comanici2025gemini}}. These models process videos in a single forward pass with large context windows and serve as strong upper-bound references, but they are not agentic frameworks and do not perform iterative exploration.

\textbf{Open-source MLLMs.} Among open-source models, we distinguish two sub-categories. The first comprises \textit{standard single-pass models} that directly answer queries from uniformly sampled frames without iterative exploration, including LongVA-7B\textcolor{myblue}{\cite{zhang2024long}}, Video-R1-7B\textcolor{myblue}{\cite{feng2026video}}, VideoChat-R1-7B\textcolor{myblue}{\cite{li2025videochatrl}}, VideoRFT-7B\textcolor{myblue}{\cite{wang2026videorft}}, Open-o3-Video-7B\textcolor{myblue}{\cite{meng2025open}}, and our base model Qwen2.5-VL-7B\textcolor{myblue}{\cite{Bai2025Qwen25VLTR}}. The second comprises \textit{agentic frameworks} that iteratively explore the video through multi-turn reasoning or tool use, including Video-MTR-7B\textcolor{myblue}{\cite{xie2025video}}, VideoMind-7B\textcolor{myblue}{\cite{liu2025videomind}}, Video-CoM-7B\textcolor{myblue}{\cite{rasheed2025video}}, LongVT-7B\textcolor{myblue}{\cite{yang2026longvt}}, and VideoTemp-o3-7B\textcolor{myblue}{\cite{liu2026videotemp}}. To ensure a fair comparison, we additionally construct an agentic variant of our base model, denoted Qwen2.5-VL-7B$^{\ddagger}$, which adopts the same multi-turn exploration framework as VideoScout but without our two-stage training. The \textbf{Agentic Framework} column in Tab.~{\ref{tab:main_results_1}} and Tab.~{\ref{tab:main_results_2}} indicates whether each method performs agentic exploration (\cmark) or operates in a single pass (\xmark).

\begin{table*}[t]
\centering
\caption{Quantitative results on long video understanding and reasoning benchmarks. The best score is in \textbf{bold}, the second-best is \underline{underlined}, and our method is highlighted with gray shading. $\ddagger$ denotes an agentic variant of the base model that adopts the same multi-turn exploration framework as VideoScout but without our two-stage training.}
\label{tab:main_results_1}
\setlength{\tabcolsep}{10.5pt}
\begin{tabular}{
l|
c| c c c c c c
}
\toprule
\textbf{Methods} &
\textbf{Agentic} &
\textbf{LVbench} &
\textbf{MINERVA} &
\textbf{MLVU}\textit{(test)} &
\multicolumn{2}{c}{{\textbf{Video-MME}}\textit{(w/o subtitle)}} &
\textbf{Average}  \\
 &
 \textbf{Framework}&
 Overall&
 Overall&
 Overall&
Overall &
Long &
 \textbf{Score}
 \\
\midrule
\multicolumn{8}{c}{\textit{Proprietary MLLMs}} \\
\midrule
GPT-4o\textcolor{myblue}{\cite{hurst2024gpt}}        & \xmark& 30.8 & 45.5 & 54.9 & 71.9 & 65.3 & 50.8  \\
Gemini-1.5-pro\textcolor{myblue}{\cite{team2024gemini}} & \xmark& 33.1 & - & - & 75.0 & 67.4 & - \\
Gemini-2.5-Pro\textcolor{myblue}{\cite{comanici2025gemini}}  & \xmark & 67.4 & 61.8 & 79.6 & 82.4 & 77.6  & 72.8 \\
\midrule
\multicolumn{8}{c}{\textit{Open-Source MLLMs}} \\
\midrule
LongVA-7B\textcolor{myblue}{\cite{zhang2024long}}    & \xmark  & 37.9 & - & 41.4  & 52.6 & 46.2 & - \\
Video-R1-7B\textcolor{myblue}{\cite{feng2026video}}    & \xmark  & 40.1 & 29.1 & 48.0  & 59.3 & 50.3 & 44.1 \\
VideoChat-R1-7B\textcolor{myblue}{\cite{li2025videochatrl}}    & \xmark  & 39.3 & 30.3 & 45.2  & 58.2 & 50.7 & 43.3 \\
VideoRFT-7B\textcolor{myblue}{\cite{wang2026videorft}}      & \xmark  & 18.7 & 29.2 & -  & 59.8 & 50.7 & - \\
Video-MTR-7B\textcolor{myblue}{\cite{xie2025video}}      & \cmark  & 42.3 & - & \underline{50.4}  & - & - & - \\
VideoMind-7B\textcolor{myblue}{\cite{liu2025videomind}}             & \cmark   & 40.8 & - & -  & 58.2 & 49.2 & - \\
Video-CoM-7B\textcolor{myblue}{\cite{rasheed2025video}}     & \cmark   & 42.5 & \underline{31.7} & 44.0  & 59.4 & - & \underline{44.4} \\
LongVT-7B\textcolor{myblue}{\cite{yang2026longvt}}             & \cmark   & 41.3 & 28.7 & 38.5  & 54.8 & 47.6 & 40.8 \\
Open-o3-Video-7B\textcolor{myblue}{\cite{meng2025open}}        & \xmark    & \underline{44.2} & 31.2 & 36.1  & \textbf{63.6} & \textbf{54.9} & 43.8\\
VideoTemp-o3-7B\textcolor{myblue}{\cite{liu2026videotemp}}          & \cmark   & 39.2 & 31.3 & 48.2  & 56.0 & 47.9 & 43.7 \\
Qwen2.5-VL-7B\textcolor{myblue}{\cite{Bai2025Qwen25VLTR}}          & \xmark   & 33.7 & 29.5 & 45.5  & 58.4 & 50.2 & 41.8 \\
Qwen2.5-VL-7B$^{\ddagger}$\textcolor{myblue}{\cite{Bai2025Qwen25VLTR}}          & \cmark   & 31.4 & 22.9 & 33.1  & 44.9 & 39.7 & 33.1 \\
\midrule
\gcell{\textbf{VideoScout-7B(Ours)}}  & \gcell{\cmark} & \gcell{\textbf{45.1}}   & \gcell{\textbf{35.4}}      & \gcell{\textbf{51.0}}   & \gcell{\underline{60.4}}   & \gcell{\underline{52.1}}   & \gcell{\textbf{48.0}}   \\
\gcell{\textit{$\Delta$ vs. Qwen2.5-VL-7B}}  & \gcell{\xmark} & \gcell{\textcolor{green!60!black}{$\uparrow$}\ 11.4} & \gcell{\textcolor{green!60!black}{$\uparrow$}\ 5.9} & \gcell{\textcolor{green!60!black}{$\uparrow$}\ 5.5} & \gcell{\textcolor{green!60!black}{$\uparrow$}\ 2.0} & \gcell{\textcolor{green!60!black}{$\uparrow$}\ 1.9} & \gcell{\textcolor{green!60!black}{$\uparrow$}\ 6.2}  \\
\gcell{\textit{$\Delta$ vs. Qwen2.5-VL-7B$^{\ddagger}$}}  & \gcell{\cmark} & \gcell{\textcolor{green!60!black}{$\uparrow$}\ 13.7} & \gcell{\textcolor{green!60!black}{$\uparrow$}\ 12.5} & \gcell{\textcolor{green!60!black}{$\uparrow$}\ 17.9} & \gcell{\textcolor{green!60!black}{$\uparrow$}\ 15.5} & \gcell{\textcolor{green!60!black}{$\uparrow$}\ 12.4} & \gcell{\textcolor{green!60!black}{$\uparrow$}\ 14.9}  \\
\bottomrule
\end{tabular}
\end{table*}

\begin{table*}[t]
\centering
\caption{Quantitative results on cross-domain video understanding benchmarks. The best score is in \textbf{bold}, the second-best is \underline{underlined}, and our method is highlighted with gray shading. $\ddagger$ denotes an agentic variant of the base model that adopts the same multi-turn exploration framework as VideoScout but without our two-stage training.}
\label{tab:main_results_2}
\setlength{\tabcolsep}{2pt}
\begin{tabular}{
l|
c| c c c c c c c c c
}
\toprule
\textbf{Methods} &
\textbf{Agentic} &
\multicolumn{4}{c}{\textbf{VideoMMMU}} &
\textbf{SciVideoBench} &
\textbf{WorldSense} &
\textbf{VSI-Bench} &
\textbf{MMVU}\textit{(mc)} &
\textbf{Average}  \\
 &
 \textbf{Framework}&
Comprehension&
Adaptation&
Perception&
Overall&
Overall&
Overall&
Overall &
Overall&
 \textbf{Score}
 \\
\midrule
\multicolumn{11}{c}{\textit{Proprietary MLLMs}} \\
\midrule
GPT-4o\textcolor{myblue}{\cite{hurst2024gpt}}        & \xmark& 62.0 & 55.6 & 66.0 & 61.2 & 24.9 & 42.6& 34.0& 71.5&46.8  \\
Gemini-1.5-pro\textcolor{myblue}{\cite{team2024gemini}} & \xmark& 53.3 & 49.3 & 59.0 & 53.8 & 27.5 & 48.0 & 42.1 & - &- \\
Gemini-2.5-Pro\textcolor{myblue}{\cite{comanici2025gemini}}  & \xmark & - & - & - & 83.6 & 64.3  & 65.1& 45.1 & 75.7 & 66.8 \\
\midrule
\multicolumn{11}{c}{\textit{Open-Source MLLMs}} \\
\midrule
LongVA-7B\textcolor{myblue}{\cite{zhang2024long}}    & \xmark  & - & - & -  & 24.0 & 14.3 & - &29.2&-&-\\
Video-R1-7B\textcolor{myblue}{\cite{feng2026video}}    & \xmark  & 46.0 & \underline{44.2} & 62.3  & 50.8 & 26.8 & 35.5 &31.8&63.8&41.7\\
VideoChat-R1-7B\textcolor{myblue}{\cite{li2025videochatrl}}    & \xmark  & 45.6 & 40.2 & 53.0  & 46.3 & 26.5 & -&33.9&64.8&- \\
VideoRFT-7B\textcolor{myblue}{\cite{wang2026videorft}}      & \xmark  & 42.0 & 36.7 & 53.1  & 44.0 & 25.7 & 38.2&-&\underline{66.7}&- \\
Video-CoM-7B\textcolor{myblue}{\cite{rasheed2025video}}     & \cmark   & - & - & -  & 50.2 & 27.6 & - &30.0&65.4&-\\
LongVT-7B\textcolor{myblue}{\cite{yang2026longvt}}            & \cmark   & 43.7 & 35.7 & 56.7  & 45.3 & 25.3 & 36.1&34.4 &63.4&40.9\\
Open-o3-Video-7B\textcolor{myblue}{\cite{meng2025open}}        & \xmark    & \underline{52.0} & 37.2 & \textbf{67.7}  & \underline{52.3} & 26.0 & 37.5&\textbf{38.5}&\textbf{67.0}&\underline{44.3}\\
VideoTemp-o3-7B\textcolor{myblue}{\cite{liu2026videotemp}}          & \cmark   & 47.0 & 43.5 & 57.3  & 49.3 & \underline{27.7} & \underline{38.7} &\underline{36.8}&64.8&43.5\\
Qwen2.5-VL-7B\textcolor{myblue}{\cite{Bai2025Qwen25VLTR}}          & \xmark   & 36.1 & 35.9 & 57.6  & 43.2 & 16.4 & 35.5&32.5&61.3&37.8 \\
Qwen2.5-VL-7B$^{\ddagger}$\textcolor{myblue}{\cite{Bai2025Qwen25VLTR}}          & \cmark   & 38.0 & 37.3 & 45.2  & 40.2 & 21.6 & 30.6&21.1&58.0&34.3 \\
\midrule
\gcell{\textbf{VideoScout-7B(Ours)}}  & \gcell{\cmark} & \gcell{\textbf{52.7}}   & \gcell{\textbf{45.5}}      & \gcell{\underline{65.9}}   & \gcell{\textbf{54.6}}   & \gcell{\textbf{28.9}}   & \gcell{\textbf{40.8}}& \gcell{35.0} & \gcell{{65.7}}& \gcell{\textbf{45.0}}  \\
\gcell{\textit{$\Delta$ vs. Qwen2.5-VL-7B}}  & \gcell{\xmark} & \gcell{\textcolor{green!60!black}{$\uparrow$}\ 16.6} & \gcell{\textcolor{green!60!black}{$\uparrow$}\ 9.6} & \gcell{\textcolor{green!60!black}{$\uparrow$}\ 8.3} & \gcell{\textcolor{green!60!black}{$\uparrow$}\ 11.4} & \gcell{\textcolor{green!60!black}{$\uparrow$}\ 12.5} & \gcell{\textcolor{green!60!black}{$\uparrow$}\ 5.3}&\gcell{\textcolor{green!60!black}{$\uparrow$}\ 2.5} & \gcell{\textcolor{green!60!black}{$\uparrow$}\ 4.4} & \gcell{\textcolor{green!60!black}{$\uparrow$}\ 7.2}  \\
\gcell{\textit{$\Delta$ vs. Qwen2.5-VL-7B$^{\ddagger}$}}  & \gcell{\cmark} & \gcell{\textcolor{green!60!black}{$\uparrow$}\ 14.7} & \gcell{\textcolor{green!60!black}{$\uparrow$}\ 8.2} & \gcell{\textcolor{green!60!black}{$\uparrow$}\ 20.7} & \gcell{\textcolor{green!60!black}{$\uparrow$}\ 14.4} & \gcell{\textcolor{green!60!black}{$\uparrow$}\ 7.3} & \gcell{\textcolor{green!60!black}{$\uparrow$}\ 10.2} & \gcell{\textcolor{green!60!black}{$\uparrow$}\ 13.9} & \gcell{\textcolor{green!60!black}{$\uparrow$}\ 7.7} & \gcell{\textcolor{green!60!black}{$\uparrow$}\ 10.7} \\
\bottomrule
\end{tabular}
\end{table*}

\begin{table*}[t]
\centering
\caption{Average number of observed frames and reasoning turns of VideoScout on each benchmark.}
\label{tab:efficiency_stats}
\setlength{\tabcolsep}{2.5pt}
\begin{tabular}{
l|
c c c c c c c c c
}
\toprule &
\textbf{LVbench} &
\textbf{MINERVA } &
\textbf{MLVU}\textit{(test)} &
{\textbf{Video-MME}}\textit{(w/o subtitle)}&
\textbf{VideoMMMU}&
\textbf{SciVideoBench}&
\textbf{WorldSense}&
\textbf{VSI-Bench}&
\textbf{MMVU}
\\
 &
 Overall&
 Overall&
 Overall&
 Overall&
 Overall&
 Overall&
 Overall&
 Overall&
 Overall
 \\
\midrule
Avg Frames     &257.4	&146.1	&129.0	&135.8&112.0&119.7&66.1&74.4&46.7  \\
Avg Turns     &16.6	&8.5	&8.1	&8.2&6.3&7.2&3.8&4.3&2.7 \\ 
\bottomrule
\end{tabular}
\end{table*}

\subsection{Main Results}

Tables~\textcolor{myblue}{\ref{tab:main_results_1}} and~\textcolor{myblue}{\ref{tab:main_results_2}} report the performance of VideoScout against proprietary and open-source baselines, and Table~\textcolor{myblue}{\ref{tab:efficiency_stats}} summarizes the average number of observed frames and reasoning turns of VideoScout on each benchmark, reflecting how its adaptive reasoning pacing allocates its observations and reasoning turns across videos of different characteristics. For a fair comparison, the number of frames used by other methods on each benchmark is also kept within a range that does not significantly exceed that of our method. VideoScout achieves the best performance among open-source methods on both benchmark categories, reaching an average of 48.0 on long video understanding and reasoning and 45.0 on cross-domain video understanding. VideoScout places first or second on most benchmarks, and its active exploration paradigm delivers steady improvements across a wide range of video types.

\textit{1) Long video understanding and reasoning.} As shown in Table~\textcolor{myblue}{\ref{tab:main_results_1}}, VideoScout obtains the best open-source results on LVBench, MINERVA, and MLVU-test, scoring 45.1, 35.4, and 51.0 respectively, and reaches the highest average score of 48.0. On Video-MME it also attains a strong second-best result of 60.4.

Compared with the two variants of our base model, VideoScout shows clear advantages in overall performance. Relative to Qwen2.5-VL-7B, which processes videos through uniform sampling, VideoScout raises the average score by 6.2 points, with per-benchmark gains of 11.4 on LVBench, 5.9 on MINERVA, 5.5 on MLVU, and 2.0 and 1.9 on the two Video-MME splits. The agentic variant Qwen2.5-VL-7B$^{\ddagger}$, which adopts the same multi-turn framework as VideoScout but without our two-stage training, performs even worse than the base model, with its average dropping from 41.8 to 33.1. This is because the untrained model tends to reason and commit to an answer prematurely rather than gathering sufficient evidence, which underscores the necessity of our two-stage training. VideoScout surpasses this untrained variant by 14.9 points on average. When further compared with other agentic frameworks such as Video-MTR-7B and VideoTemp-o3-7B, VideoScout again leads on every reported metric. These results show that adaptive reasoning pacing offers a more effective exploration strategy for locating decisive evidence in long videos than fixed multi-turn reasoning or coarse-to-fine zooming. This adaptivity is further reflected in Table~\textcolor{myblue}{\ref{tab:efficiency_stats}}: VideoScout spends more frames and reasoning turns on longer benchmarks such as LVBench, while using far fewer on shorter ones, demonstrating that it allocates its observations and reasoning turns according to video length rather than following a fixed observation schedule.

\textit{2) Cross-Domain Video Understanding.} As shown in Table~\textcolor{myblue}{\ref{tab:main_results_2}}, VideoScout obtains the highest average score of 45.0 among open-source methods. It ranks first on VideoMMMU, SciVideoBench, and WorldSense, scoring 54.6, 28.9, and 40.8 respectively, and also leads the comprehension and adaptation sub-tasks of VideoMMMU. On the remaining benchmarks it stays competitive, reaching 65.9 on the VideoMMMU perception sub-task, second only to Open-o3-Video-7B, and obtaining 35.0 on VSI-Bench and 65.7 on MMVU-mc, the latter close to the strongest open-source results. Its leading average therefore reflects strong and balanced generalization across knowledge domains rather than specialization on a few tasks.

The comparison with the two variants of our base model again highlights the importance of training. VideoScout raises the average score by 7.2 points over Qwen2.5-VL-7B and by 10.7 points over the untrained agentic variant Qwen2.5-VL-7B$^{\ddagger}$, with the latter once again falling below the base model. This reinforces our finding that the improvements arise from the two-stage training rather than the agentic framework itself. Among agentic baselines, VideoScout also surpasses the average of VideoTemp-o3-7B, which reaches 43.4, while leading on knowledge-intensive benchmarks such as VideoMMMU and SciVideoBench. Overall, these results demonstrate that adaptive exploration not only benefits long video understanding and reasoning but also transfers well to diverse cross-domain settings.

\subsection{Ablation Study}

In this section, we conduct ablation studies on the core components of VideoScout to validate the effectiveness of the two-stage training pipeline, the composite reward, and the adaptive exploration design. All ablations are evaluated on LVBench, VideoMMMU, MLVU-test, and MINERVA.

\begin{table}[t]
\centering
\caption{Ablation on the two-stage training pipeline. The best score is in \textbf{bold}, and our full model is highlighted with gray shading.}
\label{tab:training_ablation}
\setlength{\tabcolsep}{3pt}
\begin{tabular}{
l|
c c c c 
}
\toprule
\textbf{Setting} &
\textbf{LVbench} &
\textbf{VideoMMMU} &
\textbf{MLVU}\textit{(test)} &
\textbf{MINERVA}   \\
 &
 Overall&
 Overall&
 Overall&
Overall 
 \\
\midrule
SFT only     &43.4	&49.4	&47.0	&32.5  \\
RL only     &18.5	&37.7	&23.7	&8.5 \\
SFT+RL(GRPO)      &44.5	&47.1	&48.2	&31.3 \\
\gcell{\textbf{SFT+RL(DAPO)}} &
\gcell{\textbf{45.1}} &
\gcell{\textbf{54.6}} &
\gcell{\textbf{51.0}} &
\gcell{\textbf{35.4}} \\
\bottomrule
\end{tabular}
\end{table}

\textit{1) Effectiveness of the Two-Stage Training Pipeline.} We first examine the contribution of each training stage, with results reported in Table~\textcolor{myblue}{\ref{tab:training_ablation}}. Using SFT alone already provides a reasonable starting point, reaching 43.4 on LVBench, 49.4 on VideoMMMU, 47.0 on MLVU, and 32.5 on MINERVA, as cold-start fine-tuning teaches the model the multi-turn exploration format and transfers the teacher's exploration behavior. In contrast, applying RL without SFT initialization causes a severe collapse across all benchmarks, dropping to 18.5 on LVBench, 37.7 on VideoMMMU, 23.7 on MLVU, and 8.5 on MINERVA. This is because, without a cold start, the agent has no prior understanding of the action space and the output format, so it rarely produces well-formed trajectories and the sparse trajectory-level reward cannot provide a usable learning signal from scratch. These two observations indicate that the supervised and reinforcement stages are complementary and both indispensable: SFT establishes the basic exploration behavior, while RL refines it toward more accurate and efficient trajectories. We further compare two RL algorithms applied on top of the same SFT initialization. SFT+RL with GRPO brings only a marginal gain on LVBench, from 43.4 to 44.5, and even degrades on VideoMMMU and MINERVA, falling to 47.1 and 31.3. We attribute this instability to the symmetric clipping in GRPO, which couples the advantage estimation across the long multi-turn trajectories and makes the policy update sensitive to a few high-variance samples. In comparison, SFT+RL with DAPO consistently achieves the best results on all four benchmarks, reaching 45.1, 54.6, 51.0, and 35.4, and improving over SFT alone by 1.7, 5.2, 4.0, and 2.9 points respectively. These results show that the decoupled clipping bounds of DAPO yield more stable trajectory-level updates than GRPO and allow the agent to effectively refine its pace-selection and answer-timing behavior, which is essential for realizing the full potential of the two-stage training.

\begin{table}[t]
\centering
\caption{Ablation on the composite reward components in the reinforcement learning stage. The best score is in \textbf{bold}, and our full model is highlighted with gray shading.}
\label{tab:reward_ablation}
\setlength{\tabcolsep}{0.8pt}
\begin{tabular}{
l|
c c c c 
}
\toprule
\textbf{Setting} &
\textbf{LVbench} &
\textbf{VideoMMMU} &
\textbf{MLVU}\textit{(test)} &
\textbf{MINERVA}   \\
 &
 Overall&
 Overall&
 Overall&
Overall 
 \\
\midrule
RL w/o Timing Reward     &44.3	&52.4	&46.2	&33.3  \\
RL w/o Accuracy Reward     &44.4	&53.0	&48.0	&32.3 \\
RL w/o Format Reward      &42.3	&50.6	&48.4	&34.0 \\
\gcell{\textbf{VideoScout-7B(Ours)}}&
\gcell{\textbf{45.1}} &
\gcell{\textbf{54.6}} &
\gcell{\textbf{51.0}} &
\gcell{\textbf{35.4}} \\
\bottomrule
\end{tabular}
\end{table}

\begin{figure}[!t]
\centering
\includegraphics[width=3.3in]{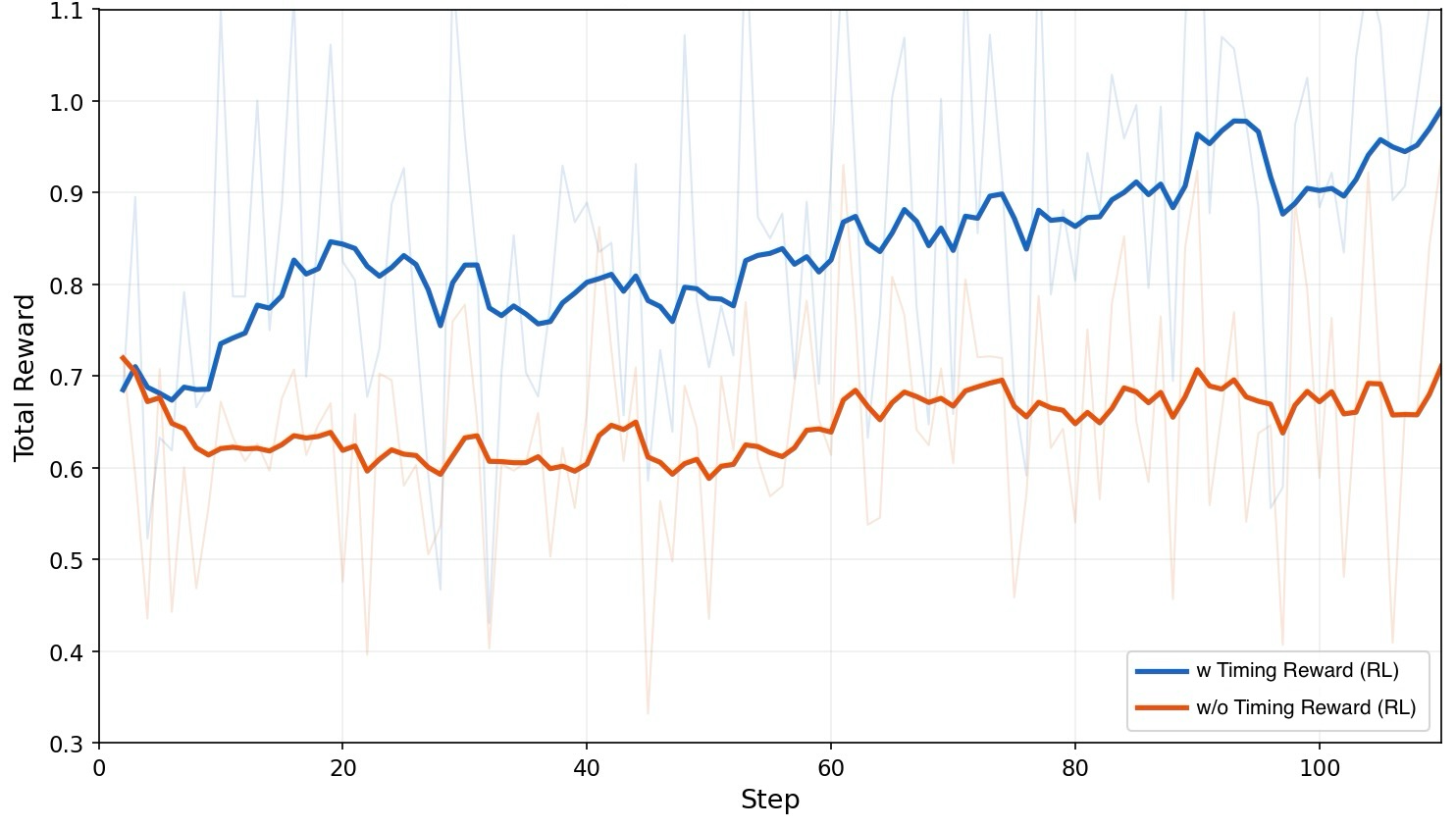}
\caption{Total reward curves during the RL stage with and without the timing reward. For clarity, we apply an exponential moving average with a smoothing coefficient of 0.9 to suppress step-wise noise and reveal the overall training trend.}
\label{fig_reward_curve}
\end{figure}

\textit{2) Effectiveness of the Composite Reward.} We further dissect the contribution of each reward component in the RL stage, as detailed in Table~\textcolor{myblue}{\ref{tab:reward_ablation}}. Removing the format reward leads to the largest degradation on LVBench, lowering the score from 45.1 to 42.3, because well-formed outputs with valid answer tags are essential for stable training and reliable answer extraction. Removing the accuracy reward also weakens performance, as it provides the primary signal for answer correctness. Removing the timing reward causes a consistent drop as well, indicating that aligning the agent's answer timing with the teacher encourages it to commit only after gathering sufficient evidence, neither prematurely nor after redundant observation. We further visualize the effect of the timing reward in Fig.~\ref{fig_reward_curve}, which plots the total reward curves during RL training with and without the timing reward. With the timing reward, the total reward rises faster and the agent gradually learns to commit to an answer at the right moment, leading to more pronounced overall improvement. Without the timing reward, the total reward barely changes, indicating that the agent fails to explore more effective behaviors and the training makes little progress. This shows that the timing reward guides the agent to answer at a more appropriate moment in the exploration process, which in turn drives more effective exploration. Overall, these results show that the three reward signals are complementary, jointly regularizing answer quality, output validity, and exploration efficiency, and that their combination is necessary for the full performance of VideoScout.

\begin{table}[t]
\centering
\caption{Ablation on the core design choices of VideoScout. The best score is in \textbf{bold}, and our full model is highlighted with gray shading. $\dagger$ denotes a setting that is re-trained with RL under the corresponding configuration, used to test whether allocating more frames brings further improvement.}
\label{tab:ablation}
\setlength{\tabcolsep}{0.1pt}
\begin{tabular}{
l|l|
c c c  
}
\toprule
\textbf{Ablation Dimension} &
\textbf{Setting} &
\textbf{LVbench}&
\textbf{VideoMMMU} &
\textbf{MLVU}\textit{(test)}   \\
 &&
 Overall&
 Overall&
 Overall
 \\
\midrule
 \multirow{3}{*}{Frame Budget}    & More frames$^{\dagger}$	&44.3	&51.8	&\textbf{51.0}  \\
    & Fewer frames	&44.8	&51.7	&48.2 \\
      & Uniform Res	&\textbf{45.3}	&54.5	&50.0 \\
\midrule
 \multirow{2}{*}{Chunk Duration}    & Fixed \SI{10}{\second}	&42.2	&48.0	&46.6  \\
    & Fixed \SI{30}{\second}	&44.5	&52.3	&49.2 \\
\midrule
 \multirow{2}{*}{ Speed Policy}    & Only $1\times$	&43.6	&49.1	&47.4  \\
    & Only $4\times$	&41.7	&50.9	&45.6 \\
\midrule
 \multirow{2}{*}{ Core Mechanism}    & w/o Notes	&42.9	&51.4	&42.0  \\
    & w/o Rewatch	&42.5	&51.6	&49.0 \\
\midrule
\gcell{\textbf{VideoScout-7B(Ours)}}&\gcell{\textbf{}}&
\gcell{45.1} &
\gcell{\textbf{54.6}} &
\gcell{\textbf{51.0}}  \\
\bottomrule
\end{tabular}
\end{table}

\textit{3) Effectiveness of the Adaptive Exploration Design.} Finally, we ablate the core design choices that enable adaptive exploration, covering the frame budget, the chunk duration, the speed policy, and the memory mechanism, with results reported in Table~\textcolor{myblue}{\ref{tab:ablation}}.

\textbf{Frame Budget.} We first study the frame budget, which determines how many frames are sampled and at what resolution under each viewing speed. As summarized in Table~\textcolor{myblue}{\ref{tab:video_sampling}}, our default configuration samples 20, 16, and 12 frames at the $1\times$, $2\times$, and $4\times$ speeds, and adopts an adaptive resolution that assigns lower per-frame resolution to faster speeds. We compare it against three alternatives. The \textit{more-frames} setting increases the per-speed frame counts to 25, 20, and 16 and is re-trained with RL under its own configuration rather than merely evaluated, so as to fairly test whether observing more frames is beneficial. The \textit{fewer-frames} setting instead reduces the counts to 12, 10, and 8. The \textit{uniform-resolution} setting keeps the same frame counts but applies the highest resolution to all speeds. As reported in Table~\textcolor{myblue}{\ref{tab:ablation}}, none of the three alternatives improves over the default budget consistently across the three benchmarks. The more-frames setting reaches only 44.3 on LVBench, still below the default score of 45.1, even though it is given additional RL training. The fewer-frames setting drops to 48.2 on MLVU, and the uniform-resolution setting marginally surpasses the default on LVBench at 45.3 versus 45.1 but falls behind on VideoMMMU and MLVU. These results suggest that increasing the number of frames mainly introduces redundant content that dilutes the truly informative evidence, whereas reducing frames means losing fine-grained temporal and spatial cues. Applying a uniformly high resolution also fails to help, since the extra visual tokens are spent on fast-forward frames that are meant for coarse coverage rather than fine inspection, yielding little accuracy gain. Overall, the default configuration achieves the best balance between information sufficiency and visual redundancy, and the advantage of VideoScout stems from learning where and how carefully to look rather than from naively observing more frames.

\textbf{Chunk Duration.} We then examine the chunk duration, which controls the temporal granularity of each observation. As summarized in Table~\textcolor{myblue}{\ref{tab:chunk_strategy}}, our default scheme adaptively determines the chunk length according to the total video duration, using shorter chunks for short videos and longer chunks for long ones. We compare it against two fixed-duration alternatives, one that splits every video into 10-second chunks and another that uses 30-second chunks regardless of video length. As reported in Table~\textcolor{myblue}{\ref{tab:ablation}}, both fixed schemes consistently underperform the adaptive one. The 10-second setting suffers the most, with the LVBench score dropping sharply from 45.1 to 42.2, while the 30-second setting alleviates this degradation but still trails the default scheme. We attribute this to the mismatch between a single fixed granularity and the wide range of video lengths. Overly short chunks fragment coherent events and require far more turns to cover the whole video, so the reasoning effort directed at the question itself is overwhelmed by redundant per-turn reasoning and memory updates, whereas overly long chunks coarsen the temporal resolution and make it harder to localize decisive moments. These results show that the chunk granularity should scale with video length rather than remain fixed.

\textbf{Speed Policy.} We further analyze the speed policy, which decides how the agent distributes its observations over time. Our default policy lets the agent freely switch among the $1\times$, $2\times$, and $4\times$ speeds during exploration. We compare it against two restricted alternatives, one that forces the agent to observe the entire video at the slow $1\times$ speed and another that forces the fast $4\times$ speed throughout. As reported in Table~\textcolor{myblue}{\ref{tab:ablation}}, both single-speed variants are clearly inferior to the learned mixed strategy. The slow-only setting obtains 43.6 on LVBench, as it spends most of its reasoning on irrelevant content before reaching the decisive evidence, whereas the fast-only setting yields the lowest LVBench score of 41.7, since it skims over fine-grained details that are necessary for accurate answering. The substantial gap between the two extremes and the default policy demonstrates that the ability to slow down on critical moments while skimming through uninformative regions is essential for efficient long-video exploration.

\begin{table}[t]
\centering
\caption{Comparison of average inference time per sample (AIT, in seconds) and accuracy (Acc) of different agentic frameworks across benchmarks. The best score is in \textbf{bold}, and our method is highlighted with gray shading.}
\label{tab:inference_time}
\setlength{\tabcolsep}{1pt}
\begin{tabular}{
l|
c c c c c c c c
}
\toprule
\textbf{Methods} &
\multicolumn{2}{c}{\textbf{LVbench}} &
\multicolumn{2}{c}{\textbf{MLVU}\textit{(test)}}&
\multicolumn{2}{c}{\textbf{MINERVA}}&
\multicolumn{2}{c}{\textbf{WorldSense}}  \\
 &
AIT(s)&
 Acc&
AIT(s)&
 Acc&
AIT(s)&
 Acc&
 AIT(s)&
 Acc
 
 \\
\midrule
LongVT-7B\textcolor{myblue}{\cite{yang2026longvt}}     &\textbf{10.9}	&41.3	&\textbf{11.0}	&38.5 &\textbf{10.7}	&28.7&9.2&36.1 \\
VideoTemp-o3-7B\textcolor{myblue}{\cite{liu2026videotemp}}       &21.3	&39.2	&18.7	&48.2 &21.7	&31.3&8.9&38.7 \\
\gcell{\textbf{VideoScout-7B(Ours)}} &
\gcell{35.1} &
\gcell{\textbf{45.1}} &
\gcell{22.3} &
\gcell{\textbf{51.0}}&
\gcell{33.6} &
\gcell{\textbf{35.4}}&\gcell{\textbf{8.1}}&\gcell{\textbf{40.8}}\\
\bottomrule
\end{tabular}
\end{table}

\textbf{Core Mechanism.} Finally, we ablate the two core mechanisms that support multi-turn exploration, namely the textual Notes that carry information across turns and the rewatch action that revisits a previous chunk for a closer look. Removing the Notes causes the largest degradation, dropping MLVU from 51.0 to 42.0, because the agent can no longer accumulate query-relevant observations across turns and is forced to reason over each turn in isolation, which is especially harmful for long videos that require integrating evidence from distant segments. Removing the rewatch action also lowers accuracy, for example from 45.1 to 42.5 on LVBench, since the agent loses the ability to return to a fast-forwarded chunk and recover the fine-grained details it has missed. These results indicate that the Notes provide cross-turn information aggregation while the rewatch action enables on-demand recovery of overlooked evidence. Together, the above ablations verify that every component of our adaptive exploration design contributes to the final performance of VideoScout.

\subsection{Efficiency Analysis}

\begin{figure*}[!t]
\centering
\includegraphics[width=7.1in]{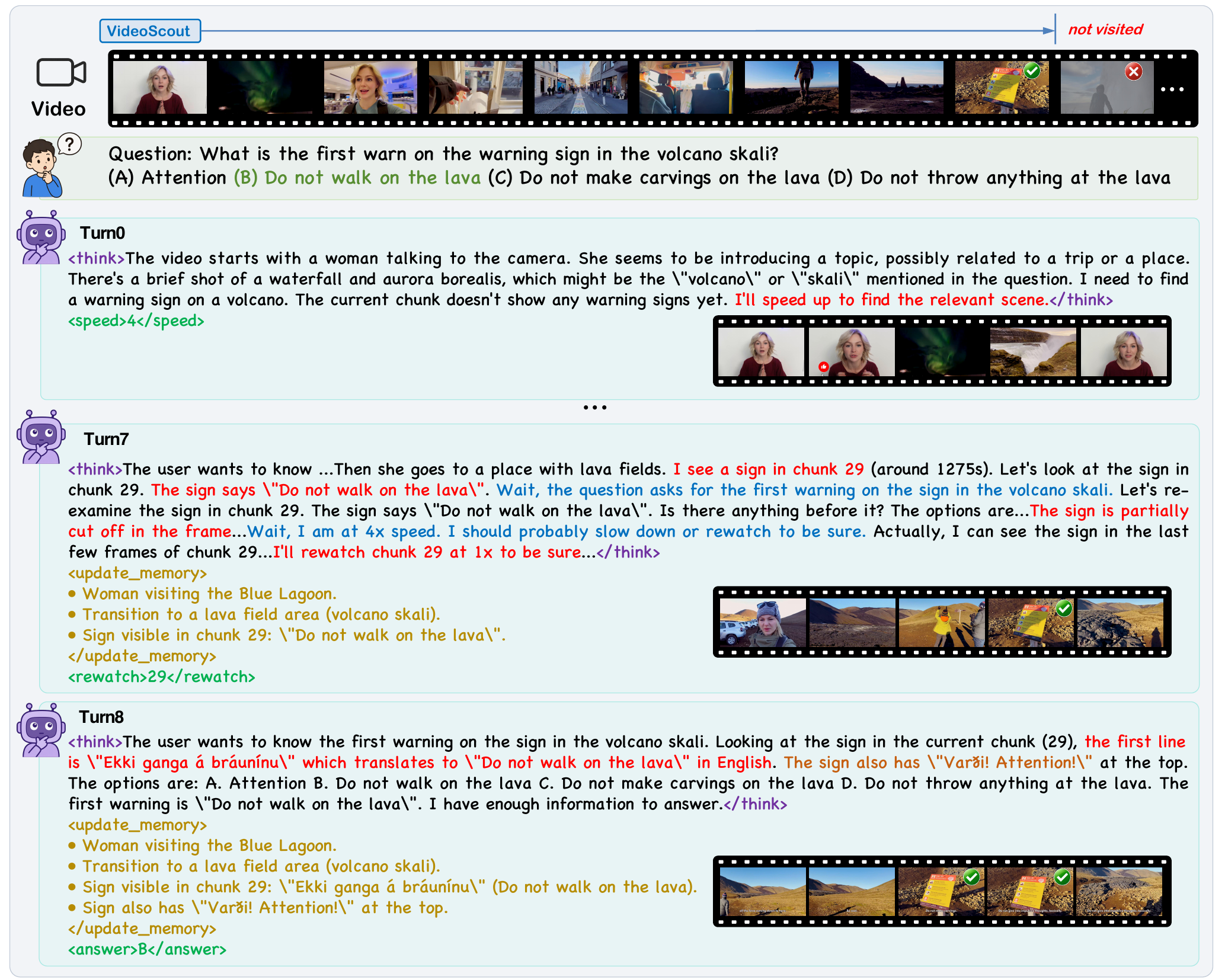}
\caption{A qualitative example of VideoScout on the LVBench dataset. Given a question about a long video, the agent reasons turn by turn: it adjusts the viewing speed, rewatches an earlier chunk when needed, updates its textual memory, and finally commits to an answer.}
\label{fig_qualita}
\end{figure*}

To investigate the inference efficiency of VideoScout, we measure the average inference time per sample (AIT) together with the accuracy on LVBench, MLVU, MINERVA, and WorldSense. All measurements are conducted on a single NVIDIA H800 GPU with a batch size of one, so that each sample is processed independently. The results are reported in Table~\textcolor{myblue}{\ref{tab:inference_time}}.

As shown in the table, VideoScout achieves the highest accuracy on all four benchmarks, and its inference time adapts substantially to the nature of each benchmark, ranging from 8.1 seconds on WorldSense to 35.1 seconds on LVBench. On WorldSense, which mainly requires short-range comprehension of everyday videos, VideoScout answers in only 8.1 seconds, which is even faster than the 8.9 seconds of VideoTemp-o3-7B and the 9.2 seconds of LongVT-7B, while still reaching the best accuracy of 40.8. On LVBench, MLVU, and MINERVA, which involve long videos and demand holistic understanding and multi-step reasoning, it instead invests more inference time, reaching 35.1, 22.3, and 33.6 seconds respectively, and this additional computation is rewarded with the highest accuracy of 45.1, 51.0, and 35.4 on the three benchmarks. By comparison, LongVT-7B and VideoTemp-o3-7B maintain a nearly fixed inference time, staying around 10.7 to 11.0 seconds for LongVT-7B and 18.7 to 21.7 seconds for VideoTemp-o3-7B across the three long-video benchmarks, since they first sample a fixed set of frames and then reason over them and can only adjust the amount of visual input through subsequent window zooming, so the observation cost is difficult to adapt to the difficulty of the query.

This behavior directly reflects the adaptive reasoning pacing of VideoScout. Rather than spending a fixed amount of observation on every video, the agent decides how long to explore according to the video length and the amount of evidence the query requires. It commits early when a short clip already provides sufficient evidence, as on WorldSense, and keeps exploring and rewatching when the answer depends on long-range or fine-grained cues, as on LVBench and MINERVA. As a result, VideoScout stays efficient on simple cases while devoting more computation to hard ones, and the higher cost it pays on difficult long videos is consistently converted into higher accuracy, instead of being a uniform and often unnecessary cost imposed on all inputs.

\subsection{Qualitative Study}

To provide an intuitive understanding of how VideoScout explores a long video, we present a qualitative example from the LVBench dataset, as shown in Fig.~\ref{fig_qualita}. The question asks for the first warning on a warning sign in the volcano Skali, which requires the agent to first locate the sign within a long video and then read its fine-grained text. In the early turns, VideoScout finds that the opening chunks contain only a person talking and some scenery unrelated to the question, so it speeds up to $4\times$ and quickly skims through these segments to gather a coarse overview while recording the visited content in its memory. After fast-forwarding to a lava-field area, it notices a warning sign in chunk 29 and reads the text as ``Do not walk on the lava''. Instead of answering immediately, the agent reflects with a ``Wait'': it realizes that it is still at $4\times$ speed, that the sign is partially cut off, and that the text may not be clear enough to determine the \emph{first} warning. It therefore actively slows down and rewatches chunk 29 at $1\times$ speed. With the higher-resolution observation, VideoScout recognizes the original line ``Ekki ganga a hrauninu'' together with a ``Vard! Attention!'' header, rules out the distractor option A, and finally commits to the correct answer B.

This example illustrates the learned ``observe--reflect--adjust'' loop of VideoScout. Rather than passively scanning at a fixed pace, the agent actively questions the sufficiency of its current observation and reflects with a ``Wait'' when the evidence is uncertain, then adjusts its viewing speed and rewatches the critical segment to obtain reliable evidence. Such reflective behavior allows VideoScout to direct its limited observations precisely toward the decisive moment and answer accurately, which is difficult to achieve with a fixed observation schedule.

\subsection{Further Analysis}

\begin{figure}[!t]
\centering
\includegraphics[width=3.3in]{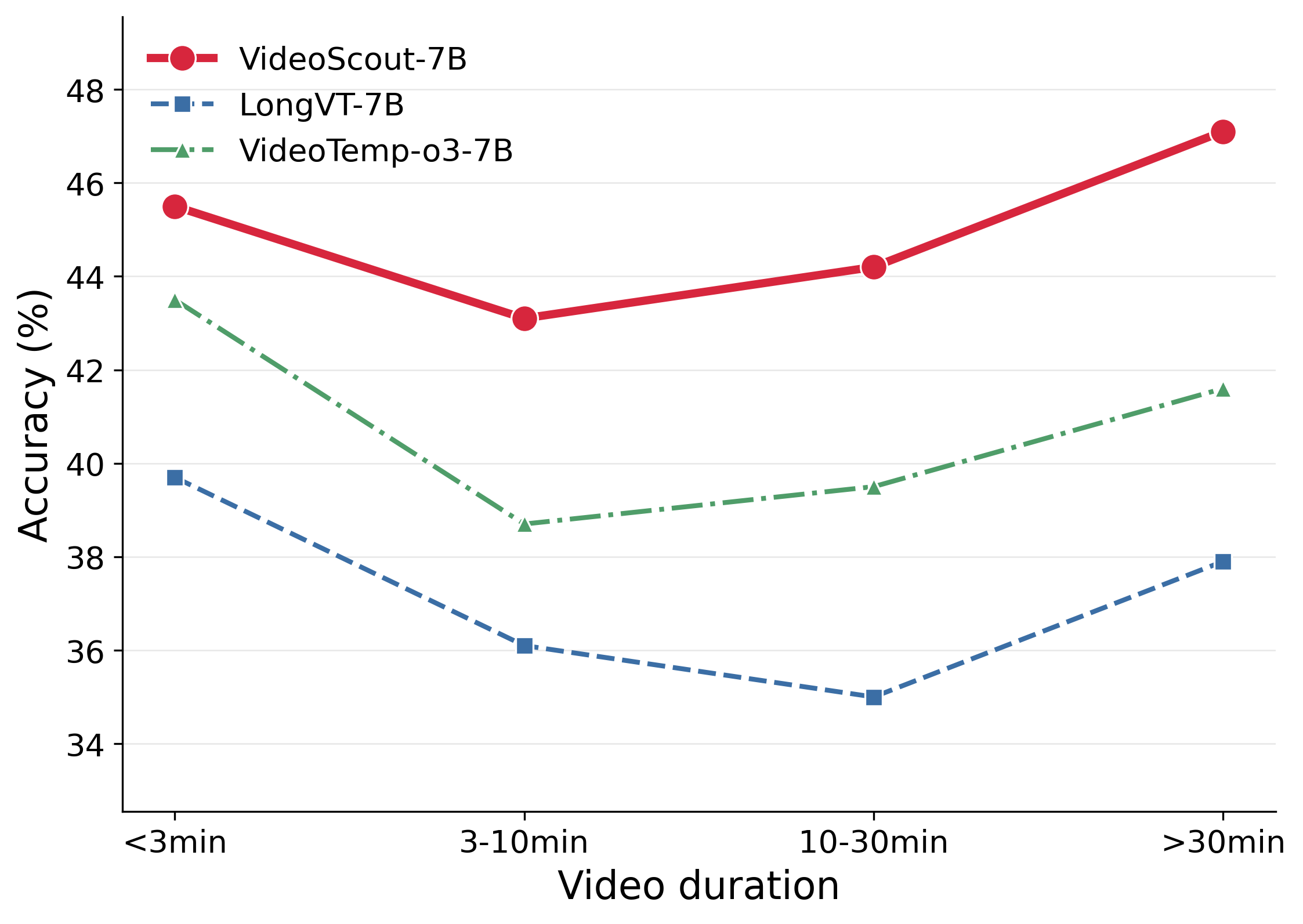}
\caption{Accuracy across video durations, averaged over all nine benchmarks pooled together. Each sample is assigned to a duration bucket according to its video length, and the accuracy within each bucket is reported for VideoScout and two agentic baselines.}
\label{fig_acc_duration}
\end{figure}

We first analyze how accuracy varies with video length. We pool the samples from all nine benchmarks, group them into four duration buckets according to video length, and report in Fig.~\textcolor{myblue}{\ref{fig_acc_duration}} the accuracy of VideoScout and two competitive agentic baselines within each bucket. VideoScout achieves the highest accuracy in every duration bucket, showing that its advantage holds consistently across all time scales rather than being confined to a particular video length. More notably, the three methods exhibit opposite trends as videos get longer. Both baselines reach their highest accuracy on the shortest videos and drop below that level once longer videos are involved, so their performance on videos longer than thirty minutes remains lower than on videos under three minutes. VideoScout instead does not degrade on long videos and even attains its best accuracy on the longest ones, where its score on videos longer than thirty minutes surpasses that on the shortest videos. This contrast indicates that coarse-to-fine zooming methods lose accuracy as the video grows longer and the decisive evidence becomes sparser, whereas VideoScout uses its adaptive reasoning pacing to spend its exploration effort on locating reliable evidence, so that its accuracy remains stable across video lengths.

\begin{figure}[!t]
\centering
\includegraphics[width=3.4in]{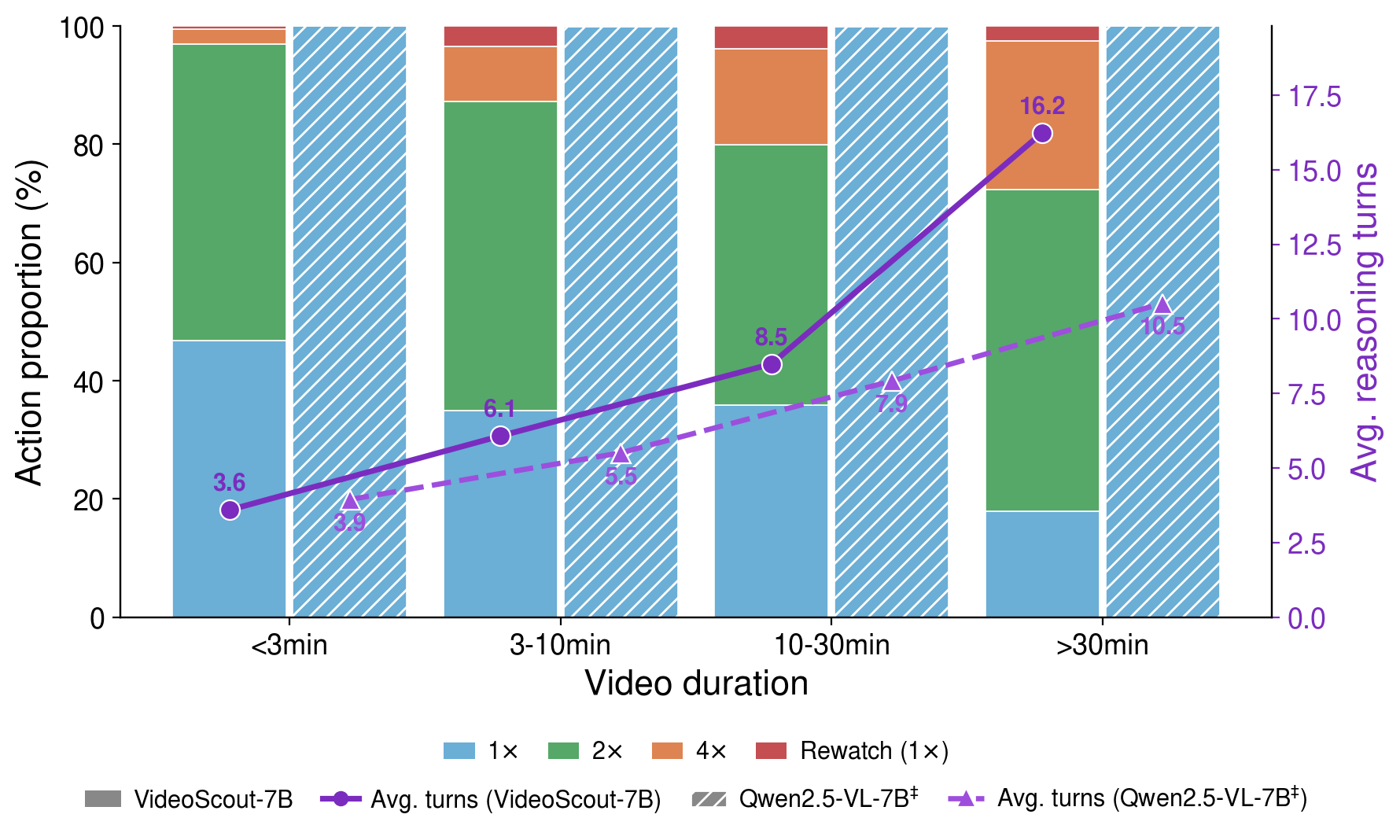}
\caption{Exploration behavior across video durations, aggregated over all nine benchmarks. The stacked bars show the proportion of each viewing action ($1{\times}$, $2{\times}$, $4{\times}$, and rewatch), and the lines report the average number of reasoning turns per sample. Solid bars and the solid line denote VideoScout, while hatched bars and the dashed line denote the untrained agentic variant Qwen2.5-VL-7B$^{\ddagger}$.}
\label{fig_behavior}
\end{figure}

To understand the behavior behind this advantage, and to verify that it is acquired through training rather than provided by the agentic framework itself, we further analyze how VideoScout adjusts its viewing pace to videos of different lengths. We group the samples from all nine benchmarks into the same four duration buckets and report in Fig.~\textcolor{myblue}{\ref{fig_behavior}} the proportion of each viewing action together with the average number of reasoning turns per sample, for both VideoScout and the untrained agentic variant Qwen2.5-VL-7B$^{\ddagger}$.

For VideoScout, the viewing-speed distribution shifts clearly with video length. On short videos, the agent watches mostly at the slow $1{\times}$ and $2{\times}$ speeds and rarely uses the fast $4{\times}$ speed. As the video grows longer, the share of $1{\times}$ steadily decreases while that of $4{\times}$ increases, so on the longest videos a large portion of the actions are issued at the fastest speed. This trend shows that VideoScout watches short videos carefully but fast-forwards more on long videos to reach informative regions, rather than inspecting every segment of a long video at full density. The average number of reasoning turns grows in parallel, rising steadily from short to long videos, which indicates that the agent takes more exploration steps when the video is longer and the evidence is harder to locate. The rewatch action appears rarely on short videos and rises to a higher proportion on medium and long videos, showing that VideoScout invokes a closer second look mainly when fast-forwarding through longer videos may have missed fine-grained evidence.

The untrained variant behaves very differently. It stays almost entirely at the $1{\times}$ speed across all duration buckets and barely uses fast-forwarding or rewatching, so its viewing pace does not adapt to video length at all. Its average number of reasoning turns is also consistently lower than that of VideoScout and grows much more slowly, indicating that it tends to commit to an answer after only a few turns instead of continuing to explore. This explains why simply placing the base model in the same multi-turn framework is not enough and even hurts performance: without our two-stage training, the agent does not learn to skim long videos or to gather sufficient evidence before answering. Taken together, these statistics confirm that VideoScout allocates its observations and reasoning turns according to video length rather than following a fixed schedule, and that this adaptive behavior is a direct result of training.

\subsection{Limitations}
\begin{figure}[!t]
\centering
\includegraphics[width=3.4in]{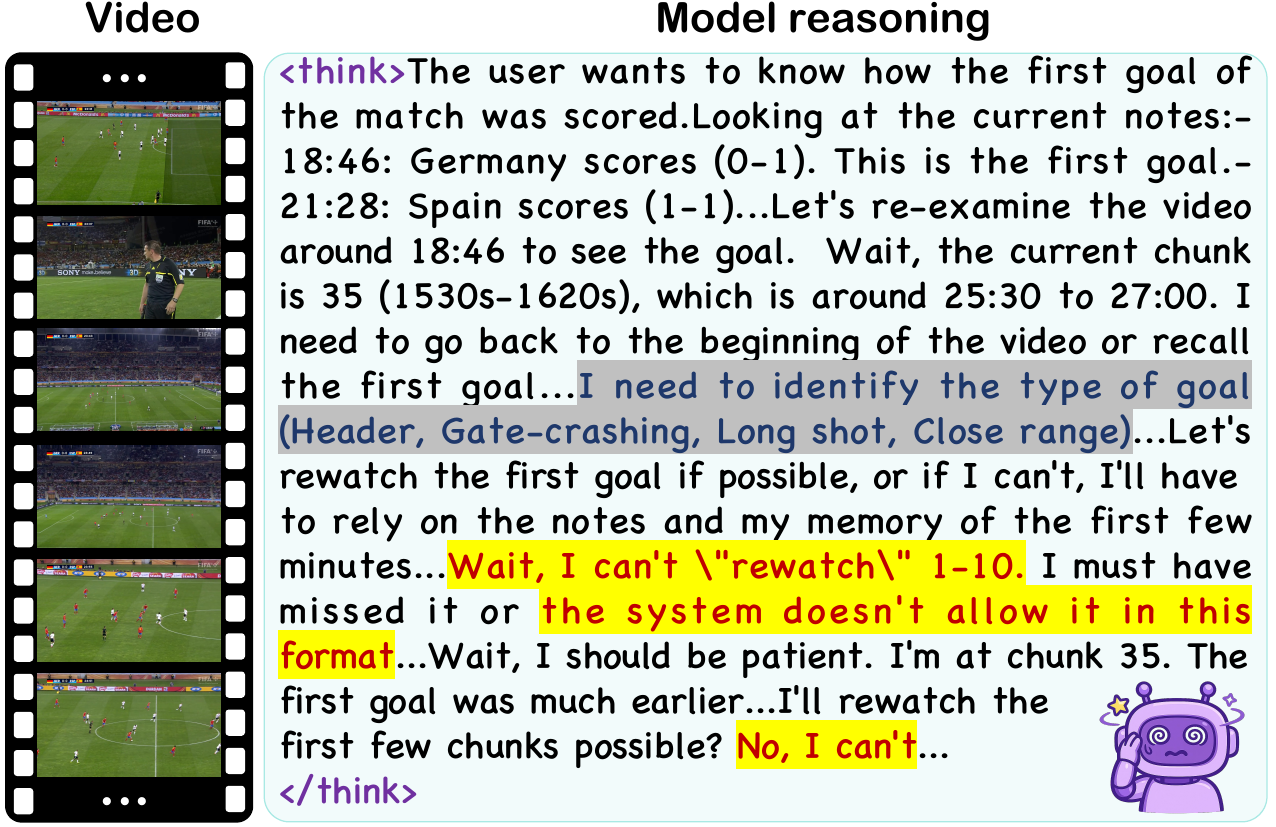}
\caption{A failure case from LVBench caused by the limited rewatch scope. The agent wants to return to early chunks for verification but cannot reach them from its current position.}
\label{fig_badcase}
\end{figure}

While VideoScout achieves strong overall performance, a few aspects remain to be improved. First, the agent carries information across turns through a textual memory that is rewritten at each update, which keeps the context compact but may also let an inaccurate observation from an early turn influence later reasoning on very long trajectories. Second, the rewatch action revisits a chunk within the most recent fast-forward window, so its scope for returning to much earlier segments is limited. As shown in Fig.~\textcolor{myblue}{\ref{fig_badcase}}, when the agent infers from its notes that the decisive evidence lies near the beginning of a long video, it is unable to rewatch those early chunks from a much later position and therefore cannot verify the evidence it needs. Third, VideoScout takes only visual input, so questions that depend on audio cues such as speech or ambient sound are beyond its present scope.

\section{Conclusion and Future Work}
In this paper, we formulated long video question answering under bounded visual context as a \textbf{Sequential Evidence Acquisition (SEA)} problem, in which an agent reads the video turn by turn along the temporal axis like a human reading a long passage. We presented VideoScout, a trained multi-turn reasoning agent that instantiates this paradigm through active exploration with adaptive reasoning pacing. To train this behavior, we constructed VideoScout-66K and adopted a two-stage pipeline that combines cold-start supervised fine-tuning with trajectory-level reinforcement learning under a composite reward over accuracy, format, and answer timing. Extensive experiments on long video understanding and reasoning benchmarks as well as cross-domain benchmarks show that VideoScout attains strong performance under a bounded per-turn visual context window while keeping inference efficient.

Building on these observations, we plan to extend VideoScout along three directions. First, we will incorporate audio perception so that the agent jointly reasons over vision and sound, enabling more complete omni-modal understanding of scenarios where the answer depends on what is heard as well as what is seen. Second, we will generalize the rewatch action into long-range revisiting, allowing the agent to return to any previously observed chunk on demand, so that evidence discovered late in the exploration can be reconciled with earlier segments. Third, we will equip the agent with a richer set of tools, such as object grounding, optical character recognition, and external knowledge retrieval, to extract fine-grained evidence and reason over information that is not directly observable from raw frames. We believe that advancing agentic exploration with adaptive reasoning pacing, by broadening its modality coverage and enriching its tool repertoire, offers a promising path toward general and efficient long video understanding.

\bibliographystyle{IEEEtran}
\bibliography{References}

\vfill

\end{document}